\documentclass[10pt,twocolumn,letterpaper]{article}

\usepackage{cvpr}              

\newcommand{\ourmethod}[1]{MAGNet}

\usepackage{graphicx} 
\usepackage{pifont}
\usepackage{multirow}
\usepackage{arydshln}

\newcommand{\cmark}{\textcolor{green!80!black}{\ding{51}}}
\newcommand{\xmark}{\textcolor{red}{\ding{55}}}

\definecolor{cvprblue}{rgb}{0.21,0.49,0.74}
\usepackage[pagebackref,breaklinks,colorlinks,allcolors=cvprblue]{hyperref}

\usepackage{booktabs} 
\usepackage{tabularx}
\usepackage{multirow}
\usepackage{wrapfig}

\def\paperID{3801} 
\def\confName{CVPR}
\def\confYear{2026}

\title{Diffusion Forcing for Multi-Agent Interaction Sequence Modeling}

\author{
Vongani H. Maluleke*\textsuperscript{$\S$} \quad
Kie Horiuchi*\textsuperscript{$\dagger$,$\S$} \quad
Lea Wilken\textsuperscript{$\S$} \quad
Evonne Ng\textsuperscript{$\ddagger$} \quad\\
Jitendra Malik\textsuperscript{$\S$} \quad
Angjoo Kanazawa\textsuperscript{$\S$} \\
\textsuperscript{$\dagger$}Sony Group Corporation\quad
\textsuperscript{$\ddagger$}Meta\quad
\textsuperscript{$\S$}UC Berkeley 
}

\begin{document}

\twocolumn[{    
    \renewcommand\twocolumn[1][]{#1}
    \maketitle
    \centering
    \includegraphics[width=1.0\linewidth]{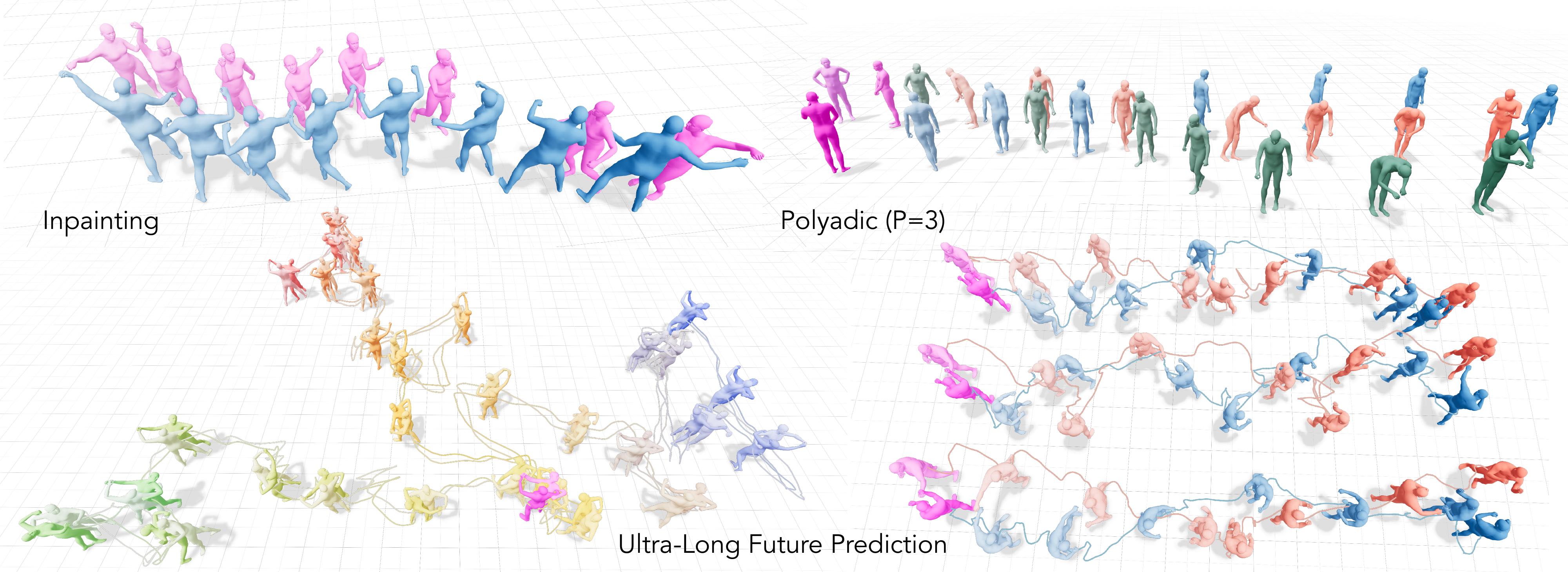}
    \captionof{figure}{\textbf{A Generative Model for Multi-Agent Interaction.} 
We propose Multi-Agent Diffusion Forcing Transformer (\ourmethod{}), a unified approach for modeling and generating realistic motion of multiple interacting humans. \ourmethod{} handles diverse interactions from synchronized activities like dancing (top-left) to arbitrary social situations (top-right) with more than two people, generating sequences that can be rolled out for hundreds of steps, with diverse samples (bottom). A single trained model supports multiple tasks at test time: Partner Inpainting (generating agent motion given complete motion of others--top left), Joint Future Prediction (predicting all agents' futures from past motions--all others), and more. The model also supports agentic (turn-taking) sampling.
\textbf{\textcolor[HTML]{FF00FF}{Pink}} indicates known conditioning poses. 
    }
    \vspace{0.5cm}
    \label{fig:teaser} 
}]

\renewcommand{\thefootnote}{\fnsymbol{footnote}}
\footnotetext[1]{Equal contribution.}
\renewcommand{\thefootnote}{\arabic{footnote}}

\begin{abstract}
Understanding and generating multi-person interactions is a fundamental challenge with broad implications for robotics and social computing. While humans naturally coordinate in groups, modeling such interactions remains difficult due to long temporal horizons, strong inter-agent dependencies, and variable group sizes. Existing motion generation methods are largely task-specific and do not generalize to flexible multi-agent generation. We introduce \ourmethod{} (Multi-Agent Generative Network), a unified autoregressive diffusion framework for multi-agent motion generation that supports a wide range of interaction tasks through flexible conditioning and sampling. \ourmethod{} performs dyadic and polyadic prediction, partner inpainting, partner prediction, and agentic generation all within a single model, and can autoregressively generate ultra-long sequences spanning hundreds of motion steps. We explicitly model inter-agent coupling during autoregressive denoising, enabling coherent coordination across agents. As a result, \ourmethod{} captures both tightly synchronized activities (\eg, dancing, boxing) and loosely structured social interactions. Our approach performs on par with specialized methods on dyadic benchmarks while naturally extending to polyadic scenarios involving three or more interacting people. Please watch the supplemental video, where the temporal dynamics and spatial coordination of generated interactions are best appreciated. \href{https://github.com/Von31/MAGNet-code}{Project Page}.

\end{abstract}
\section{Introduction}

Understanding and generating multi-person interactions is a fundamental challenge in computer vision and graphics, with applications spanning robotics, virtual reality, and social computing. Requirements are diverse: robots must react to human motion in an online streaming fashion, artists may want flexible control through keyframing, and virtual agents must produce long, natural social interactions from minimal conditioning. Moreover, social situations often extend beyond dyadic pairs to polyadic groups of three or more.

This diversity of requirements has led to a fragmented landscape of approaches. Most methods are tailored to specific dyadic tasks--- either reacting to an agent motion, or joint prediction of both agents---requiring separate models for each scenario. Architecturally, prior work either encodes inter-agent relationships through cross-attention between fixed agent pairs~\cite{ghosh2024remos, siyao2024duolando, guo2023interformer}, limiting generalization to arbitrary group sizes, or restricts inference to agentic-only sampling modes~\cite{cen2025readytoreact}, sacrificing flexibility. No single existing model supports the full spectrum of multi-agent generation tasks.

We introduce {Multi-Agent Generative Network} (\ourmethod{}), a unified framework that addresses this gap. \ourmethod{} is a transformer-based diffusion model \textit{designed} around a relational motion tokenization that flexibly supports a wide range of motion generation tasks while naturally scaling beyond two agents. 

Rather than representing motion in a global frame or as a fixed-size concatenation over agents, we construct \textbf{agent--time motion tokens}: each token corresponds to one agent at one timestep and is built to expose both \emph{what the agent does} and \emph{how it relates to others}. Concretely, a token contains a compact motion representation (via a {VQ-VAE} latent pose) together with {explicit pairwise relative transforms} to all other agents, grounded in per-frame local canonical frames. On top of this representation, we train with {diffusion forcing}~\cite{chen2025diffusion,shi2024tedi}, assigning each token an \textit{independent noise level}. This enables a single model to learn conditional distributions over \textbf{arbitrary subsets} of agents and timesteps, supporting partner prediction, inpainting, joint future prediction, and rollout within the same architecture (e.g., reactive synthesis by keeping partner tokens clean while denoising the target agent; joint rollout by keeping past tokens clean while denoising the future across agents).

Our work makes three key contributions. {First}, we design a \textbf{relational, world-agnostic} multi-agent motion representation: by embedding {pairwise transforms} in per-frame canonical coordinates, interactions are parameterized by \emph{who relates to whom} rather than absolute positions, making the model \textbf{agnostic to the number of participants} and enabling polyadic scenarios (\(P>2\)) without architectural modification. Specifically, we eliminate the reliance 
on cross-attention to inject inter-agent information. {Second}, unlike prior diffusion forcing formulation where only a single token is denoised at a time, in multi-agent generation,

agents at the same timestep must remain {coordinated} even when their tokens are corrupted with {different noise levels}. Pairing each agent's {VQ-VAE latent} with explicit pairwise transforms keeps inter-agent geometry available under any noise configuration, allowing tokens to be {denoised independently, conditioned on observed agents}. This design also enables agentic inference~\cite{cen2025readytoreact} for distributed deployment. {Third}, without any text conditioning, \ourmethod{} is competitive with specialized baselines on partner prediction and inpainting, and substantially improves the harder regimes of dyadic generation and long-horizon synthesis---\eg, improving FD by \textbf{89\%} over Ready-to-React~\cite{cen2025readytoreact}---while scaling seamlessly to polyadic interactions. Our code will be released. Please see Fig.~\ref{fig:teaser} and the supplementary video. 
\section{Related Work}

\begin{table*}[t]
\centering
\resizebox{\textwidth}{!}{
\begin{tabular}{lcccccc}
\toprule
\textbf{Method}      & \textbf{Partner Inpainting} & \textbf{Partner Prediction} & \textbf{Ultra-Long Motion} &  \textbf{Agentic Generation} & \textbf{Joint Future Prediction} & \textbf{Polyadic ($P \geq 3$)}\\
\midrule
Duolando             & \cmark & \xmark & \xmark & \xmark & \xmark & \xmark \\
ReMoS                & \cmark & \xmark & \xmark & \xmark & \xmark & \xmark \\
ReGenNet             & \cmark & \xmark & \xmark & \xmark & \xmark & \xmark \\
Human-X              & \cmark & \xmark & \xmark & \xmark & \xmark & \xmark \\
ARFlow               & \cmark & \cmark & \xmark & \xmark & \xmark & \xmark \\
Ready-to-React       & \xmark & \cmark & \cmark & \cmark & \xmark & \xmark \\
\textbf{Ours}        & \textbf{\cmark} & \textbf{\cmark} & \textbf{\cmark} & \textbf{\cmark} & \textbf{\cmark} & \textbf{\cmark} \\
\bottomrule
\end{tabular}
}

\caption{\textbf{Scope of Multi-Agent Motion Generation Methods.} We compare six capabilities: Partner Inpainting, Partner Prediction, Ultra-Long Motion, Agentic Generation, Joint Future Prediction, and Polyadic ($P\geq 3$) generation. Our approach uniquely unifies all tasks within a single model.}
\label{tab:combined_capabilities}
\end{table*}

\paragraph{Single-Agent Motion Generation.}
We focus on methods that generate the whole body motion of one or more people, for methods that predict root trajectories of pedestrians, see a recent survey~\cite{huang2025vision}. Single-agent motion generation has evolved from RNNs and VAEs ~\citep{Fragkiadaki2015RecurrentNM, julieta2017motion,aksan, maluleke2024synergy} to diffusion models \cite{dabral2023mofusion, shi2024amdm} and transformers\cite{tevet2022mdm}. While promising, these methods struggled with long-term temporal coherence and high-quality motion synthesis. Diffusion models marked a breakthrough in motion generation. Text-conditioned methods like MDM~\citep{tevet2022mdm} and MoFusion~\citep{dabral2023mofusion} generate diverse, temporally consistent human motion, while transformer-based approaches such as T2M-GPT~\citep{zhang2023t2mgpt} use self-attention to capture long-range dependencies. More recently, TEDi~\citep{shi2024tedi} advanced long sequence generation through temporally-entangled diffusion that recursively denoises a motion buffer, enabling arbitrary-length sequences without stitching artifacts. However, these methods remain limited to single-agent scenarios.

\paragraph{Dyadic Motion Generation.}
Dyadic motion generation methods model two-person interactions often by modeling inter-agent dependencies with cross-attention or diffusion. 
Text-conditioned methods like InterGen generate synchronized two-person motions from text descriptions~\citep{liang2024intergen}, while ExPI predicts future motions by modeling dependencies among agents' past trajectories~\citep{guo2022multiperson}. ReMoS and ReGenNet apply diffusion models with spatio-temporal cross-attention for partner inpainting—synthesizing one agent's motion conditioned on another's complete motion sequence~\citep{ghosh2024remos,xu2024regennet}. However, these approaches are primarily unidirectional, meaning they lack the mechanisms to treat generated outputs as reciprocal feedback that can dynamically influence other agents. ARFlow introduces a multi-modal diffusion framework for both partner inpainting and prediction, though limited to short clips~\citep{jiang2025arflow}. Human-X employs autoregressive diffusion for low-latency generation in VR/AR~\citep{ji2025humanx}, but is optimized for real-time reactive motion and cannot produce long-horizon coordinated behaviors. 
Music-conditioned approaches target dance generation. 
Duolando combines GPT architectures with off-policy reinforcement learning for music-conditioned partner inpainting~\citep{siyao2024duolando}. DuetGen and Dyadic Mamba focus on choreography-driven dance synthesis~\citep{ghosh2025duetgen,tanke2025dyadicmamba}. Methods like BUDDI~\cite{mueller2023buddi}, Reaction Priors~\cite{Fang_2024_CVPR}, and Ponimator~\cite{liu2025ponimator} learn dyadic human priors to reconstruct two people from images or video.

Closest to our approach is Ready-to-React, which unifies vector quantization, diffusion, and autoregressive generation for partner prediction in an agentic manner—each agent can independently run a model for reactive motion generation~\citep{cen2025readytoreact}. However its architecture is restricted to agentic sampling and cannot handle joint future prediction. A fundamental limitation across these methods is their inability to scale beyond dyadic interactions. The cross-attention mechanisms in Interformer~\citep{guo2023interformer}, ReMoS~\citep{ghosh2024remos}, and ReGenNet~\citep{xu2024regennet} are designed to attend from one agent to one other agent, making it unclear how to extend them when multiple other agents are present. Other methods' architectures similarly assume two-agent scenarios. Table~\ref{tab:combined_capabilities} provides a systematic comparison across five key capabilities essential for comprehensive multi-agent motion generation. Our approach can naturally handle more than two agents by adding more agent motion tokens, while unifying these diverse tasks within a single framework.
\section{Method}

We introduce the Multi-Agent Generative Network (\ourmethod{}), a unified autoregressive diffusion framework for flexible motion generation among multiple interacting agents. In our formulation, each token represents a single agent at a single timestep, and the diffusion model is trained with independent per-token noise levels via diffusion forcing~\cite{chen2025diffusion}. Each token is a composite representation encoding the agent's latent pose alongside explicit pairwise transforms to every other agent in the scene. Below we discuss how we represent the inter-agent relationships, latent motion encoding, and the Multi-Agent Generative Network.

\begin{figure}
    \centering
    \includegraphics[width=.8\linewidth]{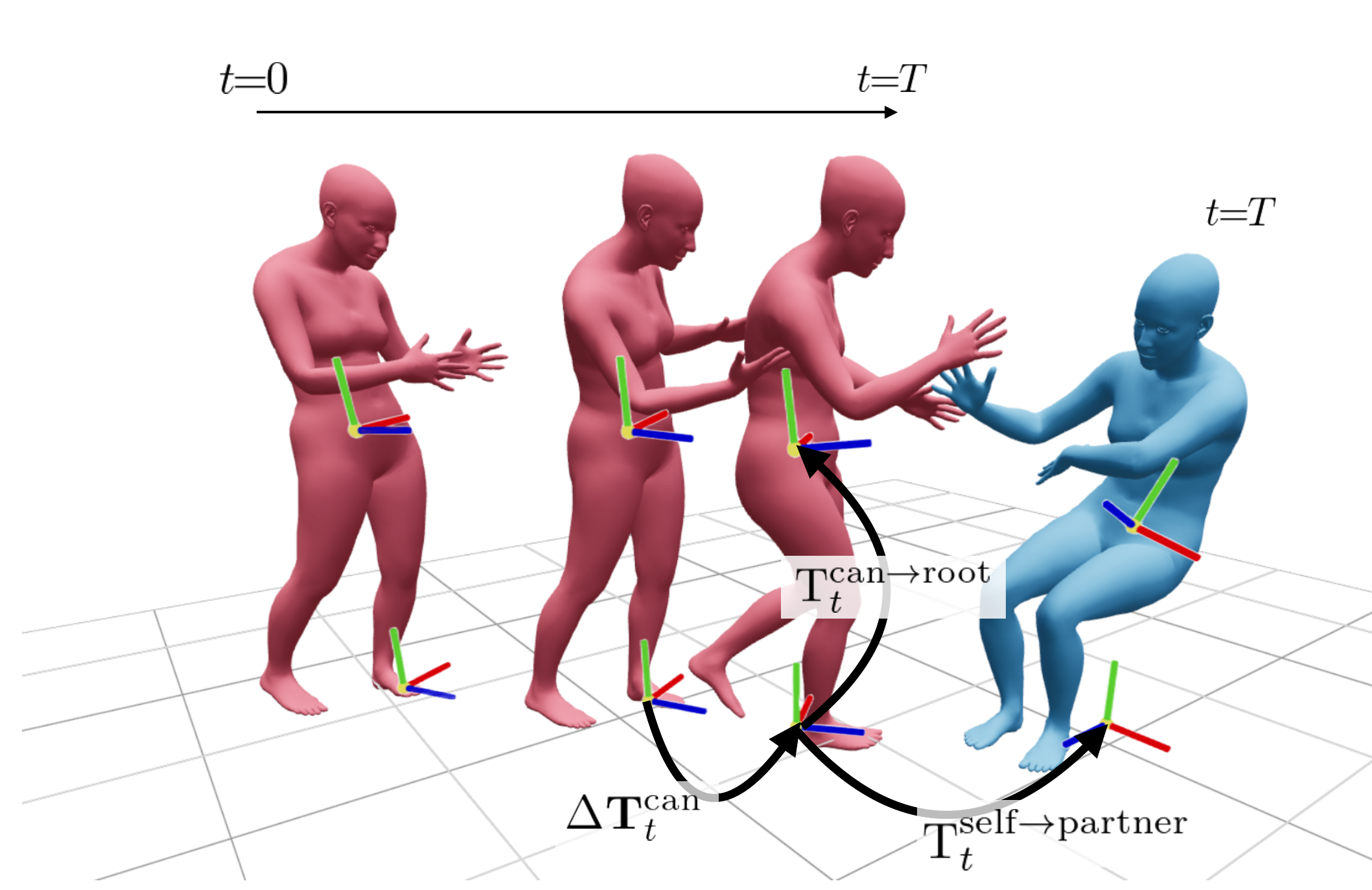}
    \caption{\textbf{Coordinate Transform Representations.} We use relative coordinate frames for both intra- and inter-person transforms, freeing the model from absolute frame definitions.}
    \label{fig:transform_repr}
\end{figure}

\begin{figure*}[t]
    \centering
    \includegraphics[width=1\linewidth]{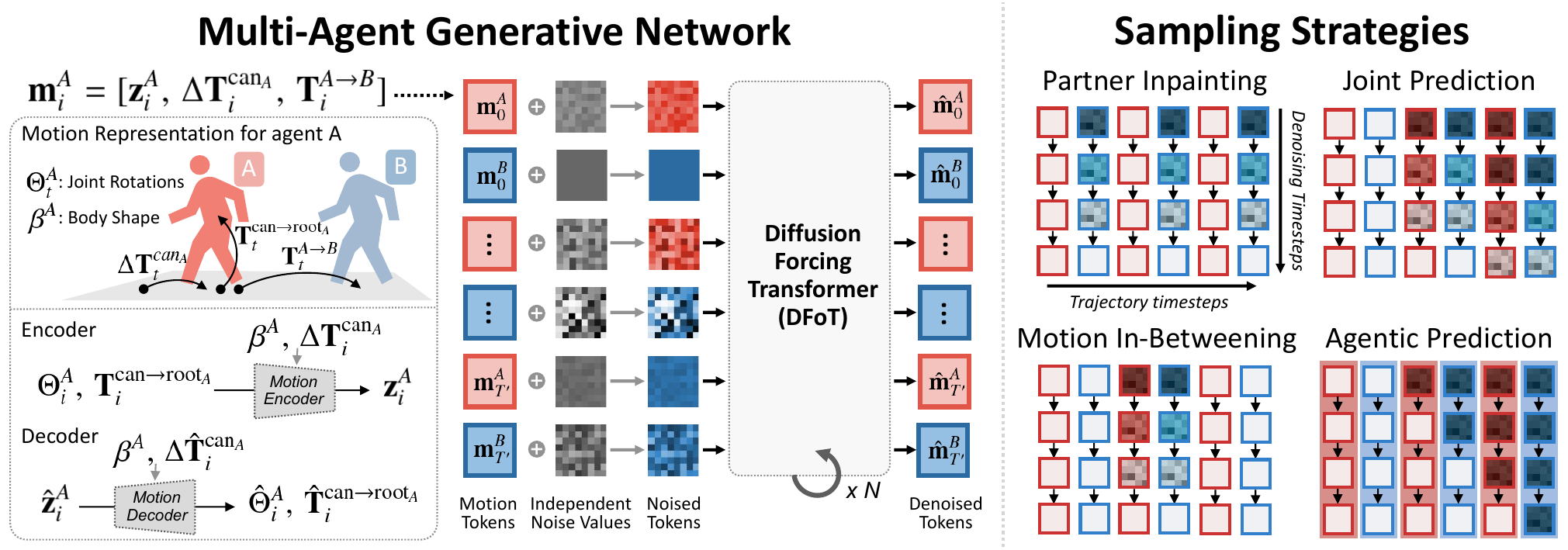}
    \caption{\textbf{ Multi-Agent Generative Network (\ourmethod{}).} \textbf{Left (Training):} Each agent’s pose at a given time is encoded by a VQ-VAE into latent pose tokens, forming motion tokens $m_i^p$ by appending latent vectors with transform parameters. Tokens from all agents across all timesteps are denoised by a transformer with independently noised tokens. \textbf{Right (Inference):} The model enables flexible conditioning: known (solid) tokens are fixed, while unknown tokens are causally denoised. This supports partner inpainting, joint prediction, and agentic turn-taking, where agents alternately generate motion and highlighted streams can run independently (\eg, on separate robots).}

    \label{fig:method}
\end{figure*}

\subsection{Motion Representation}
\label{motion_repre}

Our goal is to model a sequence of $T$ timesteps involving $P$ interacting people. Each person's motion is represented via their body shape, $\mathbf{\beta} \in \mathbb{R}^{P \times 10}$, and their joint rotations, $\Theta_t = [\theta^0_t, \dots, \theta^J_t], \Theta_t \in \mathbb{R}^{T \times P \times J \times 6}$, which can be mapped to a 3D mesh with $J$ joints using a human body model \cite{Pavlakos2019_smplifyx}. We use the 6D rotation representation for joint rotations. To capture spatial relationships while remaining invariant to absolute world positioning, we define two coordinate frames per agent: (1) a \emph{root} frame at the pelvis, and (2) a per-timestep \emph{canonical} frame obtained by projecting the root onto the ground plane, following~\cite{yi2025egoallo}, as shown in~\Cref{fig:transform_repr}. Working in canonical coordinates prevents translation magnitudes from growing unboundedly during long-horizon generation. 

We denote rigid transforms by $\mathbf{T}^{X\to Y}_t$ from $X$ frame to $Y$ frame at timestep t and parameterize each in 9D (6 for rotation, 3 for translation). In particular, the ``canonical $\to$ root transform'' at timestep $t$ is defined as:

\begin{equation}
\mathbf{T}^{\text{can} \to \text{root}}_{t} \in \mathbb{R}^{T \times P \times 9}.
\end{equation}

 Let $\Delta\mathbf{T}$ denote one-timestep transitions in a frame. The intra-agent temporal transform in the canonical frame from $t\!-\!1$ to $t$ is
\begin{equation}
\Delta\mathbf{T}^\text{can}_{t} \in \mathbb{R}^{T \times P \times 9}.
\end{equation}
 We then model pairwise inter-agent transforms at each timestep:
\begin{equation}
\mathbf{T}^{\text{self}\to\text{partner}}_{t} \in \mathbb{R}^{T \times P \times (P-1) \times 9}, 
\end{equation}
which encodes the transform from self to each partner in the canonical frame at time $t$. 
 For a scene with $P$ agents, each agent carries $P-1$ such transforms—one to every other agent—which are concatenated into the motion token representation described in Section~\ref{motion_repre}. This design allows the model to handle arbitrary group sizes without architectural modification

We adopt a latent-space approach, widely used to improve stability in long-horizon synthesis \cite{chen2023executing,rombach2021highresolution}. We first train a conditional VQ-VAE~\cite{van2017neural} restricted to the single-agent pose 

information while inter-agent and temporal relations are modeled by DFoT. 

\paragraph{VQ-VAE Encoding.}
We compress the input $x_t = (\Theta_t, \mathbf{T}^{\text{can}\to\text{root}}_t)$ into a latent representation conditioned on $c_t = (\beta, \Delta \mathbf{T}^{\text{can}}_t)$. Here, the pose $x_t$ comprises the joint rotations $\Theta_t$ and the root’s height and tilt (roll and pitch), denoted by $\mathbf{T}^{\text{can}\to\text{root}}_t$. With temporal stride $\omega=4$, the encoder yields token embeddings $\mathbf{H}=\mathrm{Enc}(x_{1:T}\mid c_{1:T})=[\,\mathbf{h}_1,\ldots,\mathbf{h}_{T/\omega}\,]$, where $\mathbf{h}_i\in\mathbb{R}^d$. The quantizer then maps each $\mathbf{h}_i$ to the nearest codebook vector, producing $\mathbf{Z} = [\mathbf{z}_1, \ldots, \mathbf{z}_{T/\omega}]$, where $\mathbf{z}_i\in\mathbb{R}^d$. Conditioning the VQ-VAE decoder is necessary because a person's body shape ($\beta$) directly influences its vertical root placement, and the temporal transform ($\Delta \mathbf{T}^\text{can}_t$) 
provides the necessary context of direction and velocity for accurate pose reconstruction.

However, we exclude $\Delta \mathbf{T}^\text{can}_t$ from the reconstruction target and instead predict it with the generative model.
This strategic division is used because $\Delta \mathbf{T}^\text{can}_t$ encodes complex temporal dynamics and is crucial for inter-agent alignment, making it structurally better suited for the sequence-modeling capabilities of the Diffusion Forcing Transformer (DFoT). By reserving the prediction of $\Delta \mathbf{T}^\text{can}_t$ (along with $\mathbf{T}^{\text{self} \to \text{partner}}$) for the DFoT, we simplify the VQ-VAE's task, allowing it to focus exclusively on learning a high-fidelity dictionary of spatial body poses ($\Theta_t$) and their corresponding per-frame root placements ($\mathbf{T}^{\text{can} \to \text{root}}_t$). 

\paragraph{VQ-VAE Loss.}
We supervise the reconstruction of joint rotations and root transformations. Throughout this section, $\tilde{\cdot}$ denotes the reconstructed values produced by the VQ-VAE decoder. We utilize a geodesic distance for rotations on $SO(3)$ and a smooth L1 (Huber) loss for translations:

\begin{equation}
\mathcal{L}_\text{VQ-VAE} 
= \lambda_{j}\sum_{j=1}^{J} d_R\!\big(\tilde{\theta}^{j}_{t},\,\theta^{j}_{t}\big) 
+ \lambda_{r}\, d_T\!\Big(\tilde{\mathbf{T}}^{\text{can}\to\text{root}}_{t},\,\mathbf{T}^{\text{can}\to\text{root}}_{t}\Big),
\end{equation}

 where the geodesic distance is $d_R(\tilde{\mathbf{R}}, \mathbf{R}) = \left\|\arccos\!\left(\tfrac{\mathrm{tr}(\tilde{\mathbf{R}}^{\!\top} \mathbf{R})-1}{2}\right)\right\|_1$, and $d_T$ combines $d_R$ over rotations with smooth L1 over translations:

\begin{equation}
d_T(\tilde{\mathbf{T}}, \mathbf{T}) = d_R(\tilde{\mathbf{R}}, \mathbf{R}) + \|\tilde{\mathbf{t}}-\mathbf{t}\|_1.
\label{eq:dt_vq_loss}
\end{equation}
 After training the VQ-VAE, we freeze its parameters and use the encoded motion for each agent as the token for modeling the joint distribution of multiple agents' motion with the DFoT discussed next.

\subsection{Multi-Agent Generative Network}
Having established our per-agent local motion representation via VQ-VAE, we now turn to modeling the joint distribution of multi-agent interactions. We train a transformer-based autoregressive diffusion model over sequences of tokens via diffusion forcing~\cite{chen2025diffusion}, where each token represents a specific agent at a specific timestep and receives an independent noise level during training. This enables flexible conditioning over motion history at inference time while preserving temporal coherence and coordination between interacting agents. Figure~\ref{fig:method} illustrates the full pipeline: \ourmethod{} operates in latent space, receiving encoded motion tokens and predicting denoised latents that are decoded back to motion trajectories.

Let $T' = T/\omega$ denote the number of token timesteps after temporal compression by the VQ-VAE. We define the motion token $\mathbf{m}_{i}^{p}$ for the $p$-th agent ($p \in \{1, \dots, P\}$) at token timestep $i$ ($i \in \{1, \dots, T'\}$) as:
\begin{equation}
\mathbf{m}_{i}^{p} = \left[\mathbf{z}_{i}^{p};\; \Delta\mathbf{T}^{\text{can}_p}_{i};\;  \mathbf{T}^{\text{self} \to \text{partner}_1}_{i};\; \dots;\; \mathbf{T}^{\text{self} \to \text{partner}_{P-1}}_{i}\right] ,
\end{equation}

where $\mathbf{m}_{i}^{p} \in \mathbb{R}^{D}$, $\mathbf{z}^p_{i}$ is the VQ-VAE latent encoding the $p$-th agent's local body motion over the window $[i\omega, (i+1)\omega)$, $\Delta\mathbf{T}^{\text{can}_p}_{i}$ is the intra-agent temporal transition over the same window, and $\mathbf{T}^{\text{self} \to \text{partner}_k}_{i}$ for $k \in \{1, \dots, P-1\}$ are the pairwise inter-agent transforms to each of the other agents. This formulation makes explicit that each token carries $P-1$ inter-agent transforms, allowing the representation to scale naturally with the number of agents. The complete set of $P \cdot T'$ motion tokens forms the sequence $\mathbf{M} \in \mathbb{R}^{(P \cdot T') \times D}$, which is processed by the DFoT.

 \paragraph{Forward Process.} 

To promote robustness and flexible conditioning, we perturb each clean latent token $\mathbf{m}$ with an independent, discrete noise level $k \in \{0, \ldots, K\}$ using a cosine schedule
$\bar{\alpha}(k) = \cos^2\left(\frac{k/K + 0.008}{1.008} \cdot \frac{\pi}{2}\right).$
The noisy token $\mathbf{m}^p_i(k^p_i)$ is then generated by applying Gaussian noise according to the sampled level $k^p_i$:
\begin{equation}
\mathbf{m}^p_i(k^p_i) = \sqrt{\bar{\alpha}(k^p_i)}\mathbf{m}^p_i + \sqrt{1 - \bar{\alpha}(k^p_i)}\boldsymbol{\epsilon}^p_i,
\end{equation}
where $\boldsymbol{\epsilon}^p_i \sim \mathcal{N}(\mathbf{0}, \mathbf{I})$. 
For each token $\mathbf{m}^{p}_{i}$, the noise level $k^p_i$ is sampled independently across tokens (i.e., i.i.d.\ over $(i, p)$).

\paragraph{Transformer Denoiser.} The Transformer denoiser $f_{\mathbf{\phi}}$ processes the noised motion token sequence

$f_{\mathbf{\phi}}$ processes the noised motion token sequence $\mathbf{M}^{(k)} = [\mathbf{m}_{1}^{1}(k_{1}^{1}), \dots, \mathbf{m}_{1}^{P}(k_{1}^{P}), \dots, \mathbf{m}^{p}_{T'}(k^{P}_{T'})] \in \mathbb{R}^{(P \cdot T') \times D}$ to predict the clean motion token sequence $\hat{\mathbf{M}}_{\text{0}}$, where each token $\mathbf{m}^{p}_{i}(k^{p}_{i})$ is perturbed by an independently sampled noise level $k^{p}_{i}$. For each token $(i, p)$, the input embedding is constructed as:
\begin{equation}
\mathbf{e}^{p}_{i} = \text{MLP}\!\left(\left[\mathbf{m}^{p}_{i}(\tau^{p}_{i});\, \text{SinEmb}(k_{i}^{p})\right]\right) + \text{RoPE}(\mathbf{m}^{p}_{i}(k_{i}^{p})) + \mathbf{\psi}(p),
\end{equation}
where $\text{SinEmb}(k^{p}_{i})$ encodes the noise level, $\text{RoPE}$~\cite{su2021roformer} injects temporal positioning, and $\mathbf{\psi}(p)$ is a learned agent-identity embedding. The transformer processes the full sequence of input embeddings $\mathbf{e}_\text{seq} = [\mathbf{e}^{1}_{1}, \dots, \mathbf{e}^{P}_{T/'}]$ to output $\hat{\mathbf{M}}_{\mathbf{0}} = f_{\mathbf{\phi}}(\mathbf{M}^{(k)}, \boldsymbol{k}_{\text{seq}})$, where $\boldsymbol{k}_{\text{seq}} = [k^{1}_{1}, \dots, k^{P}_{T'}]$ is the corresponding sequence of sampled noise levels.

\textbf{Training Objective.} \ourmethod{} is optimized using an $\mathbf{M}_{\mathbf{0}}$-prediction objective via the smooth L1 (Huber) loss across all $N = P \cdot T'$:
\begin{equation}
\mathcal{L}_{\text{Total}} = \mathbb{E}_{\mathbf{M}_{\mathbf{0}},\, \boldsymbol{k}_{\text{seq}},\, \boldsymbol{\epsilon}_{\text{seq}}} 
\left[\, \|\mathbf{M}_{\mathbf{0}} - f_{\boldsymbol{\phi}}(\mathbf{M}^{(k)}, \boldsymbol{k}_{\text{seq}})\|_1 \,\right].
\end{equation}

This loss decomposes across individual token components with weighting coefficients $\lambda$:
\begin{equation}
\mathcal{L}_{\text{Total}} = \lambda_{0}\,\mathcal{L}_{\text{pose}} +\lambda_{1}\,\mathcal{L}_{\text{cont}} +  \lambda_{2}\,\mathcal{L}_{\text{inter}} + \lambda_{3}\,\mathcal{L}_{\text{c}},
\end{equation}
where $\mathcal{L}_{\text{pose}} = \|\mathbf{Z} - \hat{\mathbf{Z}}\|_1$ supervises latent pose reconstruction, $\mathcal{L}_{\text{cont}} = \|\Delta\mathbf{T}^{\text{can}} - \Delta\hat{\mathbf{T}}^{\text{can}}\|_1$ enforces temporal continuity, and $\mathcal{L}_{\text{inter}} = \|\mathbf{T}^{\text{self}\to\text{partner}} - \hat{\mathbf{T}}^{\text{self}\to\text{partner}}\|_1$ models inter-agent spatial relationships, all defined over the full sequence. The consistency loss $\mathcal{L}_{\text{c}}$ enforces interpersonal velocity consistency by penalizing deviations between the predicted pairwise transform and its kinematically propagated counterpart:

\begin{equation}
\mathcal{L}_{\text{c}} = d_T\!\Big(\hat{\mathbf{T}}^{\text{self}\to\text{partner}},\; \big(\Delta\hat{\mathbf{T}}^{\text{self}}\big)^{-1} \hat{\mathbf{T}}_{-1}^{\text{self}\to\text{partner}}\, \Delta\hat{\mathbf{T}}^{\text{partner}}\Big),
\end{equation}

 where $d_T$ is the combined rotation-translation distance (Eq.~\ref{eq:dt_vq_loss}), $\hat{\mathbf{T}}_{-1}^{\text{self}\to\text{partner}}$ is the prediction at the previous token step, and $\Delta\hat{\mathbf{T}}^{\text{self}}$, $\Delta\hat{\mathbf{T}}^{\text{partner}}$ are the predicted canonical transforms for each agent.

\paragraph{Inference.} 

During inference, DFoT iteratively denoises the sequence $\mathbf{M}^{(k)}$ from $k=K \to 0$ 
using the learned denoiser $f_{\boldsymbol{\phi}}$, producing the final predicted clean token sequence $\hat{\mathbf{M}}_{\mathbf{0}}$. The final reconstructed physical motion sequence is then obtained via the VQ-VAE decoder: $\hat{\mathbf{X}} = \mathrm{Dec}(\hat{\mathbf{Z}} \mid \beta,\, \Delta\hat{\mathbf{T}}^{\text{can}})$, where $\hat{\mathbf{Z}}$ is the sequence of all predicted $\hat{\mathbf{z}}$ tokens and $\Delta\hat{\mathbf{T}}^{\text{can}}$ is the full sequence of intra-agent temporal transforms, both extracted from $\hat{\mathbf{M}}_{\mathbf{0}}$. The body shape $\beta$ is constant conditioning over time for each agent.

\subsection{Sampling Strategies}

Training with independent per-token noise levels enables a single model to subsume an entire family of generation tasks at inference. Without retraining or architectural changes, we simply define a temporal denoising schedule over the agent-time grid (Figure~\ref{fig:method}, right) tailored to each task via masking by noising. Observed tokens (context) are treated as ``clean'' and target tokens as ``noisy,'' and the model generates motion by coordinating denoising across both the temporal and agent axes. We describe each mode below.

\paragraph{Partner In-painting.}
Given the full trajectory of Agent~B, the model reconstructs or completes the motion of Agent~A  by treating $B$ as clean context and $A$ as the denoising target: $P(A_{0:L} \mid B_{0:L})$.

\paragraph{Partner Prediction.}
Both agents' past tokens are treated as clean context, and the model denoises Agent~A's future: $P(A_{t:t+L} \mid A_{0:t-1}, B_{0:t-1})$.

\paragraph{Joint Future Prediction.}
All agents' future motion is jointly generated from a single distribution, ensuring coordinated predictions: $P(A_{t:t+L}, B_{t:t+L} \mid A_{0:t-1},\\ B_{0:t-1})$. This preserves spatial and temporal correlations and naturally extends to $n \geq 2$ agents, as shown in the joint vs.\ independent noise sampling ablation in the supplementary. In this mode, target tokens are denoised in a causal auto-regressive manner, where the model iteratively extends the horizon, ensuring that each new prediction is temporally consistent with the previously denoised steps across the entire agent-time grid.

\paragraph{Agentic Motion Sampling.}
Each agent runs its own denoising conditioned on the others, enabling distributed deployment. In \emph{synchronous (parallel)} mode, all agents generate motion at time $t$ in parallel: $P(A_t \mid A_{0:t-1}, B_{0:t-1})$, $P(B_t \mid A_{0:t-1}, B_{0:t-1})$. In \emph{asynchronous (turn-taking)} mode, agents generate motion sequentially, enabling reactive behaviors: $P(A_t \mid A_{0:t-1}, B_{0:t-1})$, then $P(B_t \mid A_{0:t}, B_{0:t-1})$, where Agent~A's motion at timestep $t$ is generated first and then used as additional context for Agent~B.

\paragraph{Ultra-long Motion Generation.} We adopt an autoregressive windowed strategy for continuous sequence generation. The sequence is decomposed into overlapping segments with window size $W$ and overlap $O$, yielding a prediction stride $S = W - O$. At each iteration $k$, the model predicts a new segment $\tilde{z}_{k \cdot S : k \cdot S + W - 1} = f_\theta(z_{(k-1) \cdot S : (k \cdot S) - 1})$, conditioned on the final $O$ frames of the previously generated segment, ensuring temporal continuity and enabling arbitrarily long motion generation.
\section{Experiments}
\textbf{Evaluation Metrics.} We evaluate our model using standard metrics capturing positional accuracy, motion quality, physical plausibility, and multi-agent coordination: \textbf{Fréchet Distance (FD)} assesses distributional similarity between generated and real motions; \textbf{Diversity (DIV)} measures sample-level per-frame variance to ensure the model avoids mode collapse; \textbf{Foot Skating (FS)} quantifies foot sliding using average skating velocity on ground contact; \textbf{Interpenetration (IP)} uses capsule proxies to detect and measure penetration depth between body parts; \textbf{Motion Interaction (MI)} measures correlation between multiple agents by computing the difference between ground-truth and predicted correlations of their joint positions; \textbf{Mean Per Joint Position Error (MPJPE)} and \textbf{Mean Per Joint Velocity Error (MPJVE)} measure average Euclidean distance between predicted and ground-truth joint positions and velocities.

\begin{table}[h]
\centering
\scriptsize
\setlength{\tabcolsep}{4pt}
\resizebox{\columnwidth}{!}{%
\begin{tabular}{l l ccccccc}
\toprule
& \textbf{Method} & FD$\downarrow$ & DIV$\uparrow$ & MI$\downarrow$ & FS$\downarrow$ & IP$\downarrow$ & MPJPE$\downarrow$ & MPJVE$\downarrow$ \\
\midrule
\multirow{4}{*}{{\textbf{ReMoCap}}}
& RS                          & 42.151         & \textbf{24.394} & 0.406          & \textbf{0.303} & 0.522          & 1.772          & 0.041          \\
& NN                          & 28.284         & --              & 0.361          & 0.355          & 0.610          & 1.559          & 0.038          \\
& ReMoS~\cite{ghosh2024remos} & \textbf{0.002} & 0.000           & 0.003          & 0.469          & \textbf{0.162} & \textbf{0.026} & \textbf{0.012} \\
& \textbf{Ours}               & 0.029          & 0.028           & \textbf{0.000} & 0.513          & 0.176          & 0.074          & \textbf{0.012} \\
\midrule
\multirow{4}{*}{{\textbf{DD100}}}
& RS                                & 9.730          & \textbf{24.870} & 0.460          & 0.490          & 0.450          & 1.510          & 0.030          \\
& NN                                & 4.750          & --              & 0.300          & \textbf{0.340} & 0.690          & 1.580          & 0.040          \\
& Duolando~\cite{siyao2024duolando} & 18.180         & 0.000           & 0.170          & 1.880          & 0.560          & 1.680          & 0.070          \\
& \textbf{Ours}                     & \textbf{0.048}& 0.130           & \textbf{0.070} & 0.580          & \textbf{0.120} & \textbf{0.110} & \textbf{0.010} \\
\bottomrule
\end{tabular}%
}
\caption{\textbf{Partner in-painting evaluation} on ReMoCap and DD100. RS: random sampling from the dataset; NN: nearest-neighbor retrieval. We compare against ReMoS~\cite{ghosh2024remos} on ReMoCap and Duolando~\cite{siyao2024duolando} on DD100. Our model achieves the best or competitive scores across most metrics on both datasets.}
\label{tab:remos_duolando}

\end{table}

\begin{figure*}
    \centering
    \includegraphics[width=\textwidth]{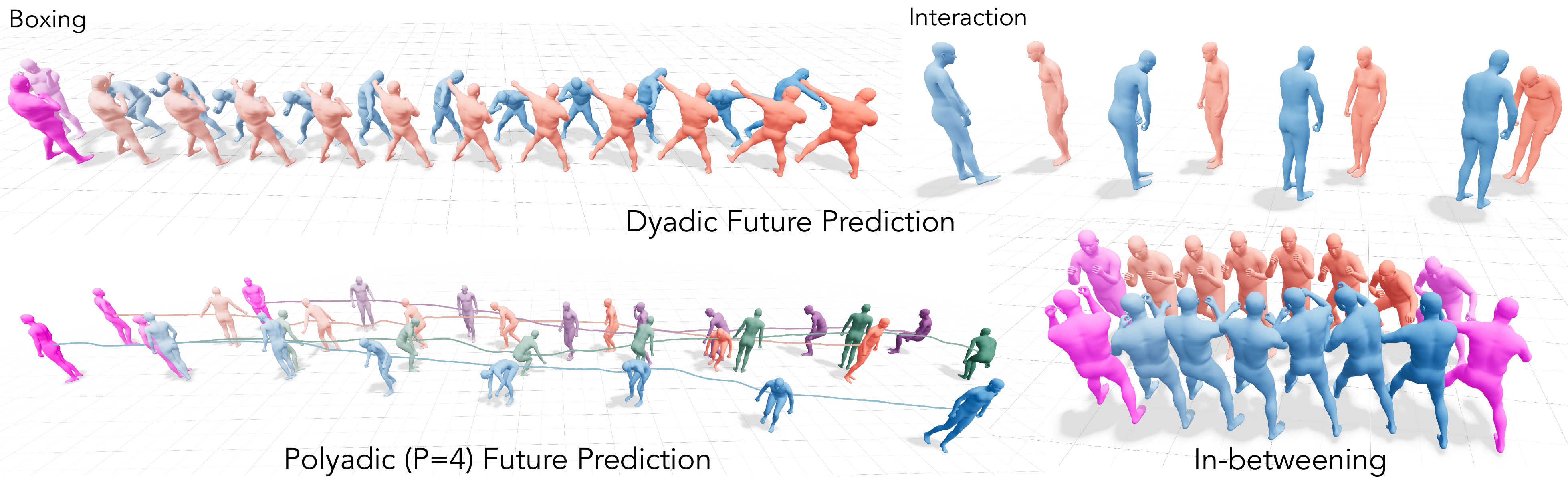}
    \caption{\textbf{Samples from our model.} We show samples from our model for different types of interaction and number of people. Our model generates realistic interactions including combat sports like boxing. In the bottom right, we show in-betweening results. 
    \textbf{\textcolor[HTML]{FF00FF}{Pink}} indicates known conditioning poses.
    Please also see the supplemental video.
    } 
    
    \label{fig:results}
\end{figure*}
\begin{figure*}[t]
    \centering
    \includegraphics[width=1\linewidth]{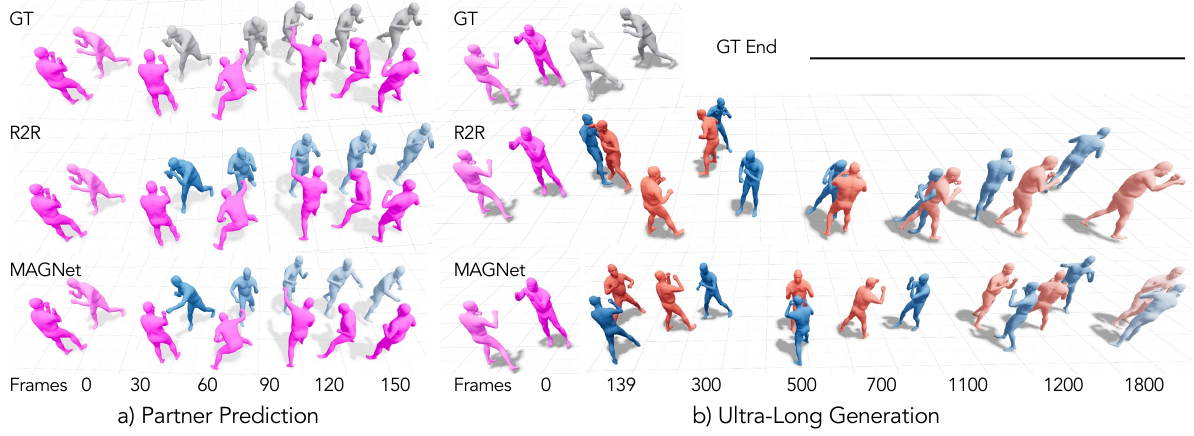}
    \caption{\textbf{Qualitative Results on Duobox.} (a) \textbf{Partner Prediction:} Given one agent's motion as context (pink), R2R and \ourmethod{} predict the partner's reactive motion (colored) against ground truth (grey, top row). Note that \ourmethod{} generates reactive counter-punches while R2R remains in a static pose. (b) \textbf{Ultra-Long Generation:} Both methods generate motion beyond the ground truth length (GT ends at frame 139, black line). R2R's agents gradually drift apart, while \ourmethod{} maintains interactive boxing across 1800 frames.
    }
    \label{fig:qual_results_pp_ul}

    \end{figure*}

\begin{figure*}[t]
    \centering
    \includegraphics[width=1\linewidth]{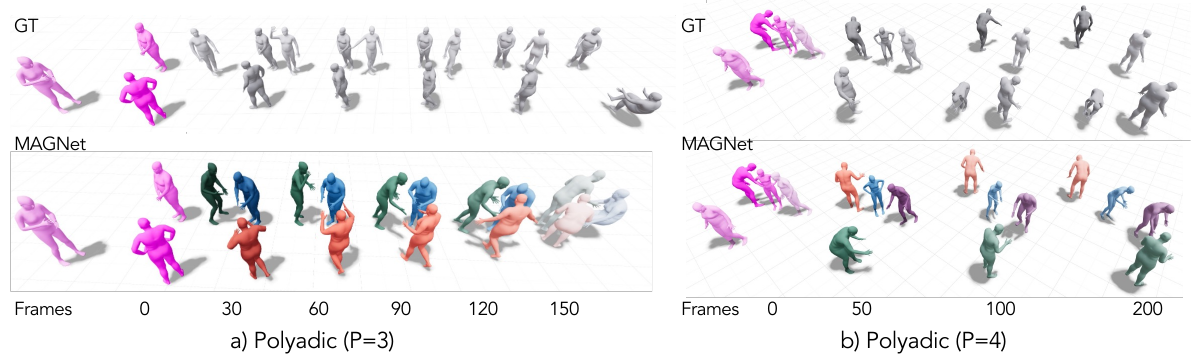}
\caption{\textbf{Polyadic Joint Future Prediction on Embody3D.} Given a context motion (pink), \ourmethod{} jointly predicts the future motions of all agents (colored, bottom row) compared to ground truth (grey, top row). (a) Three-agent prediction (P=3) over 150 frames. (b) Four-agent prediction (P=4) over 200 frames. \ourmethod{} generates physically plausible and socially coherent multi-agent motions, demonstrating its ability to scale beyond dyadic interactions.}
    \label{fig:qual_results_polyadic}
\end{figure*}

\paragraph{Data.} We evaluate our method across a diverse set of motion datasets spanning contact sports, dance, and everyday interactions. DuoBox \cite{cen2025readytoreact} captures high-contact athletic motion, while ReMoCap \cite{ghosh2024remos} and DD100 \cite{siyao2024duolando} provide complex, synchronized dance sequences. Embody 3D \cite{mclean2025embody} and Inter-X \cite{xu2024inter} represent a wide variety of multi-person social and everyday interactions. Notably, Embody 3D contains sequences involving one to four interacting individuals, while the remaining datasets focus on dyadic interactions.
In the supplemental material, we provide summary statistics for all datasets.

\paragraph{Implementation Details.} For training the 2–4 agent model on Embody 3D containing varying numbers of agents (2–4), we randomly mask $p$ agents during training.
This strategy allows the model to generalize across scenes with different agent counts without the need to train separate models for each configuration. We used Viser library to visualize our results~\cite{yi2025viser}. Please see supplemental material. for more details.

\subsection{Baselines}
\paragraph{Classic Baselines}: We consider two classic baselines: (1) Nearest Neighbor (NN), which retrieves the most similar sequence from the training set based on the motion token sequence; and (2) Random Sample, which randomly selects a sequence from the training distribution.

\paragraph{SoTA Methods}: We compare against the following state-of-the-art methods: (1) Duolando w/o music~\cite{siyao2024duolando}: An autoregressive inpainting approach that generates the follower's motion from the leader's past motion using look-ahead attention on the leader's future trajectory. We re-train Duolando without music conditioning. (2) ReMos~\cite{ghosh2024remos}: An inpainting method that generates one partner's motion conditioned on the complete future motion of the other partner. (3) Ready2React (R2R)~\cite{cen2025readytoreact}: Handles both partner prediction and dyadic motion generation. For partner prediction, it uses both agents' past motion to predict one agent's future. For dyadic prediction, it generates both agents' motion in an agentic manner from their shared motion history. We use the joint mapping provided in \cite{cen2025readytoreact} to map the MOTIVE skeleton to SMPL-X for evaluation. For fairness, we train \ourmethod{} and run inference on each dataset under similar conditions as the baseline model.

\begin{table*}[t]
\centering
\setlength{\tabcolsep}{3pt}
\resizebox{\textwidth}{!}{%
\begin{tabular}{c l ccccccc ccccccc}
\toprule
& \multirow{2}{*}{\textbf{Method}}
& \multicolumn{7}{c}{\textbf{Partner Motion Prediction}}
& \multicolumn{7}{c}{\textbf{Dyadic Motion Prediction}} \\
\cmidrule(lr){3-9} \cmidrule(lr){10-16}
& & FD$\downarrow$ & DIV$\uparrow$ & MI$\downarrow$ & FS$\downarrow$ & IP$\downarrow$ & MPJPE$\downarrow$ & MPJVE$\downarrow$
  & FD$\downarrow$ & DIV$\uparrow$ & MI$\downarrow$ & FS$\downarrow$ & IP$\downarrow$ & MPJPE$\downarrow$ & MPJVE$\downarrow$ \\
\midrule

& RS  & 0.310 & \textbf{5.176} & 0.156 & 0.277 & 0.350 & --    & --
      & 0.119 & 5.413 & 0.119 & 0.342 & 0.394 & 0.735 & 0.028 \\
& NN  & 0.766 & --    & 0.241 & \textbf{0.202} & 0.397 & 0.654 & 0.021& 0.295 & --    & 0.216 & 0.325 & 0.165 & 0.678 & 0.033 \\
& R2R~\cite{cen2025readytoreact}
      & 0.181& 0.318 & 0.071 & 0.255 & 0.309& 0.580& 0.029
      & 0.337 & 0.395 & 0.195 & 0.249 & 0.162 & \textbf{0.624} & 0.029 \\
\cmidrule(lr){2-16}
& Ours (TT: 0\%)   & -- & -- & -- & -- & -- & -- & --
                     & 0.169 & 4.809 & 0.105 & 0.342 & 0.393 & 0.685 & 0.031 \\
& Ours (TT: 50\%)  & -- & -- & -- & -- & -- & -- & --
                     & 0.495 & 1.538 & 0.104 & 0.225 & 0.249 & 0.638 & 0.028 \\
& Ours (TT: 100\%) & -- & -- & -- & -- & -- & -- & --
                     & 0.614 & 1.159 & 0.077 & \textbf{0.210} & 0.235 & 0.627 & \textbf{0.027} \\
& \textbf{Ours}
      & \textbf{0.067}& 0.669& \textbf{0.020}& 0.270& \textbf{0.019}& \textbf{0.218}& \textbf{0.016}& \textbf{0.118} & \textbf{5.622} & \textbf{0.000} & 0.407 & \textbf{0.101} & 0.714 & 0.034 \\
\bottomrule
\end{tabular}
}
\caption{\textbf{Evaluation on DuoBox.} Our model is evaluated on Partner Motion Prediction and Dyadic Motion Prediction, benchmarked against two classical baselines (RS, NN) and the SOTA method (R2R). We include an ablation on Dyadic Prediction to analyze the effect of offsetting denoising steps for synthesizing Turn-Taking (TT) variations.}
\label{tab:duobox}

\end{table*}

\subsection{Quantitative Results}
\paragraph{Partner Inpainting.} On the in-painting task in  \Cref{tab:remos_duolando}, our method performs comparably to existing approaches, with varying strengths across datasets. On ReMoCap, our results are largely on-par with the dedicated ReMoS model \cite{ghosh2024remos}: FS, IP, and MPJPE remain close (FS 0.513 vs. 0.469, IP 0.176 vs. 0.162, MPJPE 0.074 vs. 0.026). ReMoS achieves exceptionally low FD; however, this comes at the cost of notably low diversity (DIV 0.028 vs. 0.000). Our model, in contrast, achieves slightly higher FD but produces more diverse outputs, offering a more balanced trade-off between realism and variation. The NN and RS baselines achieve good results in foot sliding which is to be expected while the interaction metrics (MI and IP) and motion realism of two people does not compete with our model. On the DD100 test set, we obtain the lowest (best) FD score (0.05 vs. 18.18 for Duolando (trained without music) \cite{siyao2024duolando}) and competitive performance across other metrics. The relatively high FD score of DD100 suggests that the music signal is an important resource for Duolando. In both cases, our model outperforms previous works in terms of motion interaction indicating it's ability to model realistic human interaction.

\paragraph{Partner Prediction.} In \Cref{tab:duobox} we compare our model to \mbox{Ready-to-React \cite{cen2025readytoreact}} on the DuoBox test set. On the partner-prediction task, our method outperforms R2R on nearly all metrics, achieving the best FD (0.067 vs.\ 0.181), MI (0.020 vs.\ 0.071), IP (0.019 vs.\ 0.309), MPJPE (0.218 vs.\ 0.580), and MPJVE (0.016 vs.\ 0.029). These gains suggest that diffusion forcing over multi-agent tokens is effective even in the simpler single-agent prediction setting, where it's only conditioned on past motion.

\paragraph{Dyadic Prediction.} On the more challenging dyadic motion generation task, where both agents are generated, our model attains superior FD and DIV scores, indicating better global motion stability and diversity when generating full two-person interactions. Moreover, our approach produces the lowest Interpenetration and Foot Skating errors in the dyadic setting, underscoring the physical plausibility and interaction fidelity of our generated motions.

\paragraph{Polyadic Future Prediction.} A key advantage of our architecture is that a single trained model naturally extends to any number of agents without modification. To our knowledge, no existing learned method supports generation beyond two agents; we therefore evaluate against the only universally applicable references: NN and RS. Table~\ref{tab:polyadic} reports results on Embody3D as N increases from 2 to 4 using the same model. Our method achieves the best FD across all group sizes and maintains the lowest MPJPE and foot skating (FS), with the advantage widening at N=4—where NN and RS see sharp FD degradation due to data sparsity, our model degrades gracefully ($0.298\rightarrow 0.590$ vs.\ $0.659 \rightarrow 2.143$for NN). NN achieves low MI and MPJVE by construction, as it retrieves real sequences, but this does not extend to realistic generation for larger groups. RS yields high diversity but lacks inter-agent coherence. Our method's lower diversity relative to RS reflects a natural consequence of generating coordinated group motion rather than independent samples, as confirmed qualitatively in Figure~\ref{fig:qual_results_polyadic}.

\begin{table}[t]
\centering
\scriptsize 
\setlength{\tabcolsep}{4pt}
\begin{tabular}{l l ccccccc} 
\toprule
\textbf{N} & \textbf{Method} & FD$\downarrow$ & DIV$\uparrow$ & MI$\downarrow$ & FS$\downarrow$ & IP$\downarrow$ & MPJPE$\downarrow$ & MPJVE$\downarrow$ \\
\midrule
\multirow{3}{*}{2}
& RS   & 0.885          & \textbf{7.408}  & \textbf{0.052}          & 0.332          & \textbf{0.007} & 0.668          & 0.008 \\
& NN   & 0.659          & --              & 0.065          & 0.471          & 0.224          & 0.646          & \textbf{0.007} \\
& Ours & \textbf{0.298} & 6.219          & 0.093          & \textbf{0.207} & 0.044          & \textbf{0.561} & 0.011 \\
\cmidrule{2-9} 
\multirow{3}{*}{3}
& RS   & 0.431          & \textbf{7.948}  & 0.011          & 0.284          & \textbf{0.036} & 0.617          & \textbf{0.006} \\
& NN   & 0.566          & --              & 0.019          & 0.298          & 0.700          & 0.539          & \textbf{0.006} \\
& Ours & \textbf{0.293} & 6.313          & \textbf{0.004} & \textbf{0.257} & 0.043          & \textbf{0.526} & 0.011 \\
\cmidrule{2-9} 
\multirow{3}{*}{4}
& RS   & 2.039          & \textbf{12.945} & 0.007          & 0.310          & 0.054          & 0.894          & \textbf{0.009} \\
& NN   & 2.143          & --              & \textbf{0.001} & 0.363          & 0.573          & 0.880          & \textbf{0.009} \\
& Ours & \textbf{0.590} & 3.018          & 0.058          & \textbf{0.237} & \textbf{0.050} & \textbf{0.546} & \textbf{0.009} \\
\bottomrule
\end{tabular}
\caption{\textbf{Polyadic quantitative evaluation on Embody3D.} Best results are bolded per $N$ and per metric.}
\label{tab:polyadic}
\end{table}

\begin{figure}
  \centering

    \includegraphics[width=\linewidth]{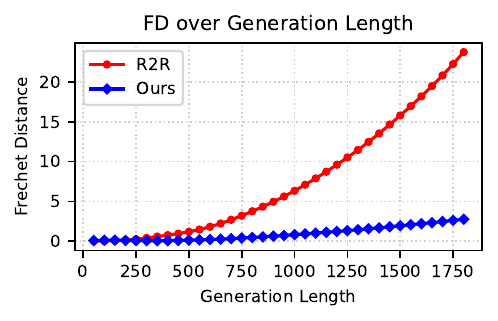}

    \caption{\small \textbf{Cumulative FD over generation length.}}
  \label{fig:cumulative_fd}

  \end{figure}
  
\paragraph{Ultra-Long Generation.} Our method significantly outperforms R2R over 1800 frames on Duobox with 4 frames as context, achieving an FD of \textbf{2.781} vs.\ 24.810, MI of \textbf{0.021} vs.\ 0.062, and IP of \textbf{0.116} vs.\ 0.345. Figure \ref{fig:cumulative_fd} shows that R2R's FD accumulates exponentially, while \ourmethod{} maintains superior stability, demonstrating robustness to temporal drift.

Overall, these results underscore the key advantage of our approach: despite not being tailored to any single task or dataset, our method remains competitive across all metrics and particularly strong on those reflecting long-horizon coherence, high realism, and interaction quality. This balance of generality and performance demonstrates the value of our unified formulation.

\paragraph{Ablations.} To assess our architectural design choices, we ablate key components and analyze their impact on generation quality in \Cref{tab:ablation}. \textbf{No VQ-VAE (Raw Joints and $\mathbf{T}^{\text{can} \to \text{root}}$ )} assesses the contribution of motion quantization, we replace the discrete VQ-VAE embeddings with continuous body parameters and canonical-to-root transformations. Table \ref{tab:ablation} shows that without VQ-VAE, the overall model performance declines significantly. 
\textbf{Importance of $\mathbf{T}^{\text{self} \to \text{partner}}_{t}$}: To evaluate the impact of explicit inter-agent spatial alignment, we remove the relative transformation $\mathbf{T}^{\text{self} \to \text{partner}}_{t}$ from the motion tokens. The removal of this inter-agent spatial encoding leads to a significant degradation in model performance. As shown in \Cref{tab:ablation}, FD and MPJPE deteriorate (rising from 0.052 to 0.126 and 0.641 to 0.665, respectively), while penetration artifacts nearly triple, increasing from 0.116 to 0.342. These results demonstrate that explicit spatial alignment modeling is essential for synthesizing physically plausible and coordinated interactions.

\begin{table}[t]
\centering
\scriptsize 
\setlength{\tabcolsep}{4pt} 
\begin{tabularx}{\columnwidth}{l ccccccc}
\toprule
\textbf{Variant} & FD$\downarrow$ & DIV$\uparrow$ & MI$\downarrow$ & FS$\downarrow$ & IP$\downarrow$ & MPJPE$\downarrow$ & MPJVE$\downarrow$ \\
\midrule
w/o VQ-VAE
  & 4.476          & 5.174           & 0.255          & 5.682          & \textbf{0.057}          & 1.128          & 0.200 \\
w/o $\mathbf{T}^{\text{self} \to \text{partner}}$
  & 0.126          & \textbf{10.761} & 0.128          & \textbf{0.367}          & 0.342          & 0.665          & \textbf{0.032}          \\
\midrule
\textbf{\ourmethod{} (Ours)}
  & \textbf{0.052} & 9.572           & \textbf{0.124} & 0.423 & 0.116 & \textbf{0.641} & 0.034 \\
\bottomrule
\end{tabularx}
\caption{the VQ-VAE causes catastrophic foot skating (FS: 5.682) and FD degradation ($86\times$), indicating that discrete tokenization regularizes motion quality. Removing $\mathbf{T}^{\text{self} \to \text{partner}}$ nearly triples interpenetration (IP: 0.116 $\to$ 0.342), confirming that explicit inter-agent spatial encoding is critical for coordination.}
\label{tab:ablation}
\end{table}

We further analyze in ~\Cref{tab:duobox}, the effect of varying the offset interval in our agentic turn-taking sampling schedule, which determines when subsequent tokens begin denoising. This ablation studies how different offset steps influence the generation. We observe that smaller offset steps lead to more dynamic and responsive interactions, whereas larger offsets promote smoother but less reactive motion, as the agent waits until the other agent is completely denoised. Please see the supplementary video for results on more inference strategies.

\subsection{Qualitative Evaluation}
Our qualitative results corroborate the quantitative findings. In Figure~\ref{fig:qual_results_pp_ul}(a), our partner predictions maintain plausible boxing exchanges over 150 frames and generate timely reactive motion such as counter-punches in response to the context agent's strikes, while R2R produces less reactive motions. Figure~\ref{fig:qual_results_pp_ul}(b) highlights the long-horizon advantage: \ourmethod{} sustains coherent two-person boxing well beyond the 139-frame ground truth (up to 1800 frames), whereas R2R's agents gradually drift apart and the interaction breaks down. Figure~\ref{fig:qual_results_polyadic} demonstrates polyadic generation on Embody3D with three and four agents using the same trained model, producing coordinated group interactions without architectural changes. Figure~\ref{fig:results} illustrates the breadth of tasks handled by a single model: boxing, casual interaction, polyadic generation (P=4), and in-betweening. We encourage viewing the supplementary video, where temporal dynamics and inter-agent coordination are best appreciated.

\section{Conclusion}

We present \ourmethod{}, a unified framework for multi-agent motion generation that subsumes dyadic and polyadic future prediction, partner inpainting, ultra-long synthesis, and agentic generation—tasks that previously required separate, specialized models.
This is enabled by two core design choices: a pairwise relative transform representation that generalizes to arbitrary numbers of agents by construction, and diffusion forcing over multi-agent tokens that implicitly learns all conditional distributions over any subset of agents and frames. Without text conditioning, \ourmethod{} is on-par and outperforms specialized baselines across multiple motion generation tasks: we improve FD by 89\% on long-horizon generation over R2R~\cite{cen2025readytoreact}, achieve the lowest FD on DD100 partner inpainting (0.05 vs.\ 18.18 for Duolando), and scale gracefully from two to four agents on Embody3D where retrieval baselines degrade. We discuss limitations in the appendix. We hope our unified formulation can serve as a foundation for broader multi-agent generation, with future work targeting swarm-scale scenarios involving hundreds of socially interacting agents.

\section{Acknowledgments}
We would like to thank Brent Yi and Chung Min Kim for their visualization contributions to this work, and Alexander Richard and David McAllister for their helpful consultations and support. This project was funded in part by Sony and Meta BAIR partners, NSF CAREER, and ONR MURI N00014-21-1-280

{
    \small
    \bibliographystyle{ieeenat_fullname}
    \bibliography{main}

@article{van2017neural,
  title={Neural discrete representation learning},
  author={Van Den Oord, Aaron and Vinyals, Oriol and others},
  journal={Advances in neural information processing systems},
  volume={30},
  year={2017}
}

@inproceedings{yi2025egoallo,
    title={Estimating body and hand motion in an ego-sensed world},
    author={Yi, Brent and Ye, Vickie and Zheng, Maya and Li, Yunqi and M{\"u}ller, Lea and Pavlakos, Georgios and Ma, Yi and Malik, Jitendra and Kanazawa, Angjoo},
    booktitle={Proceedings of the Computer Vision and Pattern Recognition Conference},
    pages={7072--7084},
    year={2025}
}

@article{chen2025diffusion,
  title={Diffusion forcing: Next-token prediction meets full-sequence diffusion},
  author={Chen, Boyuan and Mart{\'\i} Mons{\'o}, Diego and Du, Yilun and Simchowitz, Max and Tedrake, Russ and Sitzmann, Vincent},
  journal={Advances in Neural Information Processing Systems},
  volume={37},
  pages={24081--24125},
  year={2025}
}

@article{huang2025vision,
  title={Vision-Based Multi-Future Trajectory Prediction: A Survey},
  author={Huang, Renhao and Xue, Hao and Pagnucco, Maurice and Salim, Flora D and Song, Yang},
  journal={IEEE Transactions on Neural Networks and Learning Systems},
  year={2025},
  publisher={IEEE}
}

@inproceedings{liu2025ponimator,
  author    = {Liu, Shaowei and Guo, Chuan and Zhou, Bing and Wang, Jian},
  title     = {Ponimator: Unfolding Interactive Pose for Versatile Human-Human Interaction Animation},
  booktitle = {Proceedings of the IEEE/CVF International Conference on Computer Vision (ICCV)},
  year      = {2025}
}

@article{mclean2025embody,
  title={Embody 3D: A Large-scale Multimodal Motion and Behavior Dataset},
  author={McLean, Claire and Meendering, Makenzie and Swartz, Tristan and Gabbay, Orri and Olsen, Alexandra and Jacobs, Rachel and Rosen, Nicholas and de Bree, Philippe and Garcia, Tony and Merrill, Gadsden and others},
  journal={arXiv preprint arXiv:2510.16258},
  year={2025}
}

@misc{yi2025viser,
      title={Viser: Imperative, Web-based 3D Visualization in Python},
      author={Brent Yi and Chung Min Kim and Justin Kerr and Gina Wu and Rebecca Feng and Anthony Zhang and Jonas Kulhanek and Hongsuk Choi and Yi Ma and Matthew Tancik and Angjoo Kanazawa},
      year={2025},
      eprint={2507.22885},
      archivePrefix={arXiv},
      primaryClass={cs.CV},
      url={https://arxiv.org/abs/2507.22885}}

@inproceedings{mueller2023buddi,
        title={Generative Proxemics: A Prior for {3D} Social Interaction from Images},
        author={M{\"u}ller, Lea, Lea and Ye, Vickie and Pavlakos, Georgios and Black, Michael J. and Kanazawa, Angjoo},
        booktitle = {{Computer Vision and Pattern Recognition (CVPR)}},
        year={2024}}

@article{liang2024intergen,
  title={Intergen: Diffusion-based multi-human motion generation under complex interactions},
  author={Liang, Han and Zhang, Wenqian and Li, Wenxuan and Yu, Jingyi and Xu, Lan},
  journal={International Journal of Computer Vision},
  volume={132},
  number={9},
  pages={3463--3483},
  year={2024},
  publisher={Springer}
}

@INPROCEEDINGS{aksan,
  author={Aksan, Emre and Kaufmann, Manuel and Cao, Peng and Hilliges, Otmar},
  booktitle={2021 International Conference on 3D Vision (3DV)}, 
  title={A Spatio-temporal Transformer for 3D Human Motion Prediction}, 
  year={2021},
  volume={},
  number={},
  pages={565-574},
  doi={10.1109/3DV53792.2021.00066}}

@article{
        shi2024amdm,
        author = {Shi, Yi and Wang, Jingbo and Jiang, Xuekun and Lin, Bingkun and Dai, Bo and Peng, Xue Bin},
        title = {Interactive Character Control with Auto-Regressive Motion Diffusion Models},
        year = {2024},
        issue_date = {August 2024},
        publisher = {Association for Computing Machinery},
        address = {New York, NY, USA},
        volume = {43},
        journal = {ACM Trans. Graph.},
        month = {jul}
}

@inproceedings{Pavlakos2019_smplifyx,
	title        = {Expressive Body Capture: {3D} Hands, Face, and Body From a Single Image},
	author       = {Pavlakos, Georgios and Choutas, Vasileios and Ghorbani, Nima and Bolkart, Timo and Osman, Ahmed A. A. and Tzionas, Dimitrios and Black, Michael J.},
	year         = 2019,
	month        = jun,
	  booktitle      = {{Computer Vision and Pattern Recognition (CVPR)}}
}

@article{maluleke2024synergy,
    title={Synergy and Synchrony in Couple Dances},
    author={Maluleke, Vongani H and M{\"u}ller, Lea and Rajasegaran, Jathushan and Pavlakos, Georgios and Ginosar, Shiry and Kanazawa, Angjoo and Malik, Jitendra},
    journal={arXiv preprint arXiv:2409.04440},
    year={2024}
  }

@inproceedings{guo2022multiperson,
  title     = {Multi-Person Extreme Motion Prediction},
  author    = {Wen Guo and Xiaoyu Bie and Xavier Alameda-Pineda and Francesc Moreno-Noguer},
  booktitle = {CVPR},
  year      = {2022},
  note      = {arXiv:2105.08825}
}

@InProceedings{ghosh2024remos,
  title     = {ReMoS: 3D Motion-Conditioned Reaction Synthesis for Two-Person Interactions},
  author    = {Anindita Ghosh and Rishabh Dabral and Vladislav Golyanik and Christian Theobalt and Philipp Slusallek},
  booktitle = {European Conference on Computer Vision (ECCV)},
  year      = {2024}
}

@inproceedings{guo2023interformer,
  title     = {InterFormer: Interleaved Transformer for Two-Person Interactive Motion Prediction},
  author    = {Wen Guo and Francesc Moreno-Noguer and Xavier Alameda-Pineda},
  booktitle = {International Conference on Computer Vision (ICCV)},
  year      = {2023},
  note      = {arXiv:2207.01685}
}

@inproceedings{cen2025readytoreact,
  author    = {Zhi Cen and Huaijin Pi and Sida Peng and Qing Shuai and Yujun Shen and Hujun Bao and Xiaowei Zhou and Ruizhen Hu},
  title     = {Ready-to-React: Online Reaction Policy for Two-Character Interaction Generation},
  booktitle = {International Conference on Learning Representations (ICLR)},
  year      = {2025}
}

@inproceedings{xu2024regennet,
  author    = {Liang Xu and Yizhou Zhou and Yichao Yan and Xin Jin and Wenhan Zhu and Fengyun Rao and Xiaokang Yang and Wenjun Zeng},
  title     = {ReGenNet: Towards Human Action-Reaction Synthesis},
  booktitle = {CVPR},
  year      = {2024}
}

@inproceedings{siyao2024duolando,
  author    = {Li Siyao and Tianpei Gu and Zhitao Yang and Zhengyu Lin and Ziwei Liu and Henghui Ding and Lei Yang and Chen Change Loy},
  title     = {Duolando: Follower GPT with Off-Policy Reinforcement Learning for Dance Accompaniment},
  booktitle = {International Conference on Learning Representations (ICLR)},
  year      = {2024}
}

@InProceedings{Fang_2024_CVPR,
  author    = {Fang, Qi and Fan, Yinghui and Li, Yanjun and Dong, Junting and Wu, Dingwei and Zhang, Weidong and Chen, Kang},
  title     = {Capturing Closely Interacted Two-Person Motions with Reaction Priors},
  booktitle = {Proceedings of the IEEE/CVF Conference on Computer Vision and Pattern Recognition (CVPR)},
  year      = {2024},
  pages     = {655-665}
}

@inproceedings{ghosh2025duetgen,
  title     = {DuetGen: Music Driven Two-Person Dance Generation via Hierarchical Masked Modeling},
  author    = {Anindita Ghosh and Bing Zhou and Rishabh Dabral and Jian Wang and Vladislav Golyanik and Christian Theobalt and Philipp Slusallek and Chuan Guo},
  booktitle = {ACM SIGGRAPH Conference Track},
  year      = {2025},
  note      = {arXiv:2506.18680}
}

@article{tanke2025dyadicmamba,
  title     = {Dyadic Mamba: Long-term Dyadic Human Motion Synthesis},
  author    = {Julian Tanke and Takashi Shibuya and Kengo Uchida and Koichi Saito and Yuki Mitsufuji},
  journal   = {arXiv preprint arXiv:2505.09827},
  year      = {2025}
}

@inproceedings{ji2025humanx,
  title     = {Towards Immersive Human-X Interaction: A Real-Time Framework for Physically Plausible Motion Synthesis},
  author    = {Kaiyang Ji and Ye Shi and Zichen Jin and Kangyi Chen and Lan Xu and Yuexin Ma and Jingyi Yu and Jingya Wang},
  booktitle = {ICCV},
  year      = {2025},
  note      = {arXiv:2508.02106}
}

@article{jiang2025arflow,
  author    = {Wentao Jiang and Jingya Wang and Kaiyang Ji and Baoxiong Jia and Siyuan Huang and Ye Shi},
  title     = {ARFlow: Human Action-Reaction Flow Matching with Physical Guidance},
  journal   = {arXiv preprint arXiv:2503.16973},
  year      = {2025}

}

@article{tevet2022mdm,
  title     = {Human Motion Diffusion Model},
  author    = {Guy Tevet and Sigal Raab and Brian Gordon and Yonatan Shafir and Daniel Cohen-Or and Amit H. Bermano},
  journal   = {arXiv preprint arXiv:2209.14916},
  year      = {2022},
  note      = {https://arxiv.org/abs/2209.14916}
}

@inproceedings{dabral2023mofusion,
  title     = {MoFusion: A Framework for Denoising-Diffusion-based Motion Synthesis},
  author    = {Rishabh Dabral and Muhammad Hamza Mughal and Vladislav Golyanik and Christian Theobalt},
  booktitle = {Proceedings of the IEEE/CVF Conference on Computer Vision and Pattern Recognition (CVPR)},
  year      = {2023},
  note      = {https://arxiv.org/abs/2212.02837}
}

@inproceedings{zhang2023t2mgpt,
  title     = {T2M-GPT: Generating Human Motion from Textual Descriptions with Discrete Representations},
  author    = {Jianrong Zhang and Yangsong Zhang and Xiaodong Cun and Shaoli Huang and Yong Zhang and Hongwei Zhao and Hongtao Lu and Xi Shen},
  booktitle = {Proceedings of the IEEE/CVF Conference on Computer Vision and Pattern Recognition (CVPR)},
  year      = {2023},
  note      = {https://arxiv.org/abs/2301.06052}
}

@inproceedings{shi2024tedi,
  title     = {TEDi: Temporally-Entangled Diffusion for Long-Term Motion Synthesis},
  author    = {Zhi Shi and Pengfei Wan and Nguyen Nguyen and Sifei Liu and Dimitris Metaxas and Lei Yang and Yebin Liu},
  booktitle = {Proceedings of the IEEE/CVF Conference on Computer Vision and Pattern Recognition (CVPR)},
  year      = {2024}
}

@inproceedings{su2021roformer,
  title={RoFormer: Enhanced Transformer with Rotary Position Embedding},
  author={Su, Jianlin and Lu, Yu and Pan, Shengfeng and Wen, Bo and Liu, Yunfeng},
  booktitle={Proceedings of the 2021 Conference on Empirical Methods in Natural Language Processing (EMNLP)},
  year={2021},
  url={https://arxiv.org/abs/2104.09864}
}

@article{Fragkiadaki2015RecurrentNM,
  title={Recurrent Network Models for Human Dynamics},
  author={Katerina Fragkiadaki and Sergey Levine and Panna Felsen and Jitendra Malik},
  journal={2015 IEEE International Conference on Computer Vision (ICCV)},
  year={2015},
  pages={4346-4354},
  url={https://api.semanticscholar.org/CorpusID:128024}
}

@inproceedings{julieta2017motion,
  title={On human motion prediction using recurrent neural networks},
  author={Martinez, Julieta and Black, Michael J. and Romero, Javier},
  booktitle={CVPR},
  year={2017}
}

@inproceedings{xu2024inter,
  title={Inter-x: Towards versatile human-human interaction analysis},
  author={Xu, Liang and Lv, Xintao and Yan, Yichao and Jin, Xin and Wu, Shuwen and Xu, Congsheng and Liu, Yifan and Zhou, Yizhou and Rao, Fengyun and Sheng, Xingdong and others},
  booktitle={CVPR},
  pages={22260--22271},
  year={2024}
}

@inproceedings{chen2023executing,
  title     = {Executing your Commands via Motion Diffusion in Latent Space},
  author    = {Chen, Xin and Jiang, Biao and Liu, Wen and Huang, Zilong and Fu, Bin and Chen, Tao and Yu, Gang},
  booktitle = {Proceedings of the IEEE/CVF Conference on Computer Vision and Pattern Recognition},
  pages     = {18000--18010},
  year      = {2023},
}

@misc{rombach2021highresolution,
      title={High-Resolution Image Synthesis with Latent Diffusion Models}, 
      author={Robin Rombach and Andreas Blattmann and Dominik Lorenz and Patrick Esser and Björn Ommer},
      year={2021},
      eprint={2112.10752},
      archivePrefix={arXiv},
      primaryClass={cs.CV}
}

@inproceedings{dhariwal2021diffusion,
  title={Diffusion Models Beat GANs on Image Synthesis},
  author={Dhariwal, Prafulla and Nichol, Alexander},
  booktitle={Advances in Neural Information Processing Systems},
  volume={34},
  pages={8780--8794},
  year={2021},
  url={https://proceedings.neurips.cc/paper/2021/file/49ad23d1ec9fa4bd8d77d02681df5cfa-Paper.pdf}
}
}

\newpage
\clearpage
\setcounter{page}{1}
\maketitlesupplementary
\renewcommand{\thetable}{A.\arabic{table}}
\setcounter{table}{0} 
\renewcommand{\thesection}{A.\arabic{section}}
\setcounter{section}{0}  
\renewcommand{\thefigure}{A.\arabic{figure}}
\setcounter{figure}{0} 

 This is the supplementary material for our main paper ``Diffusion Forcing for Multi-Agent Interaction Sequence Modeling''. We provide details about data (\cref{sec:data_supp}) and model (\cref{sec:models_supp}). 

Our supplementary material also includes a video file demonstrating motions generated by our model across different datasets (dancing, interaction, sports), as well as ultra-long sequences and multi-person generation.

\section{Data}
\label{sec:data_supp}

We consider a diverse collection of datasets covering different types of human motion, including contact sports, dance, and day-to-day interactions. Specifically:

\begin{itemize}
    \item \textbf{Contact Sports:} We use the \textit{DuoBoX}\cite{cen2025readytoreact} dataset to represent high-contact motion sequences.
    \item \textbf{Dance:} We include several motion capture dance datasets, namely \textit{ReMoCap (LindyHop)}\cite{ghosh2024remos} and \textit{DD100} \cite{siyao2024duolando}.

    \item \textbf{Day-to-Day Interactions:} For everyday activities, we consider \textit{Embody 3D} \cite{mclean2025embody}, and \textit{Inter-X} \cite{xu2024inter}. 
\end{itemize}
 Table \ref{tab:datasets_summary} summarizes the key statistics of these datasets, including frame rates, number of clips, subjects, actions, total frames, and approximate total duration.

\begin{table*}[hbt!]
\centering
\footnotesize
\resizebox{\textwidth}{!}{
\begin{tabular}{lcccccc}
\toprule
\textbf{Dataset} & \textbf{FPS} & \textbf{Clips} & \textbf{Unique Subjects} & \textbf{Unique Actions} & \textbf{Total Frames} & \textbf{Total Time} \\
\midrule

DD100 \cite{siyao2024duolando}          & 30                & 167                 & 5                   & 10 genres     & 350.4K                   & 3.2 hrs \\
DuoBox \cite{cen2025readytoreact}             & 120               & 116                 & 3                   & 1             & 913.8K                   & 2.1 hrs\\

ReMoCap (LindyHop) \cite{ghosh2024remos}          & 50   & 8  & 4     & 1   & 174.2K   & 0.9 hrs  \\
Inter-X \cite{xu2024inter}           & 120               & 11,388              & 89                  & 40            & 8,071.8K                 & 18.7 hrs \\
Embody 3D (2 people) \cite{mclean2025embody}     & 30            & 66                  & 4                   & 24            & 577.7K                   & 5.3 hrs \\
Embody 3D (3 people) \cite{mclean2025embody}    & 30            & 35                  & 3                   & 18            & 273.0K                   & 2.5 hrs \\
Embody 3D (4 people) \cite{mclean2025embody}    & 30            & 612                 & 70                  & 56            & 5,169.7K                 & 47.9 hrs \\
\bottomrule
\end{tabular}
}
\caption{\textbf{Summary Statistics of Multi-Agent Motion Datasets}}
\label{tab:datasets_summary}

\end{table*}

\section{Limitations}
\begin{figure}
    \centering
    \includegraphics[width=0.5\linewidth]{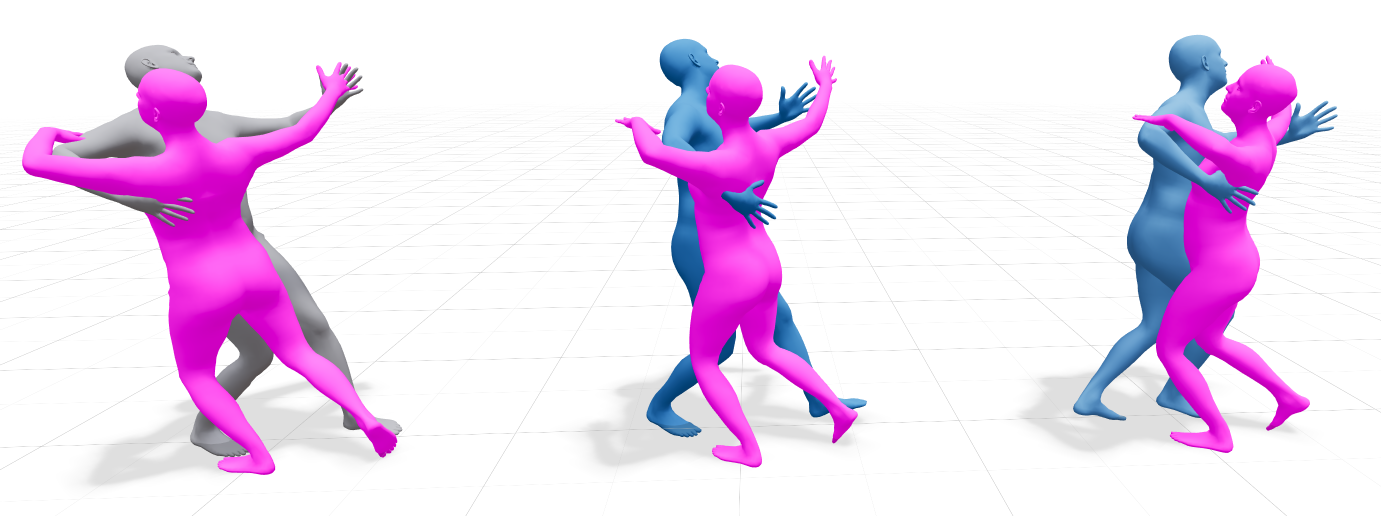}
    \caption{\textbf{Example of an inter-agent penetration artifact generated by \ourmethod{}}. Trained without explicit physical constraints, the model fails to enforce non-collision, causing Agent A’s hand to pass through Agent B’s torso during the contact. This reflects a common limitation of data-driven motion models trained solely on motion capture data}
    \label{fig:penetration}
\end{figure}

While \ourmethod{} demonstrates strong results that do not drift over long generation horizons, we occasionally observe inter-agent penetration, where one agent's limbs intersect with their partner's body mesh (Figure \ref{fig:penetration}). This issue stems from training on motion capture data without explicit physical constraints—indeed, some intersection artifacts exist in the training data itself. Our current inference scheme is simple to demonstrate the capability of the base model; incorporating guidance mechanisms to prevent penetrations are promising directions for future work. Another interesting direction is using these kinematic predictions as a controller for physics-based animation systems, which can address these issues.

\section{Model Architecture Details}
\label{sec:models_supp}

We train the VQ-VAE and DFoT models using the AdamW optimizer with an initial learning rate of $2\times10^{-4}$, weight decay of $1\times10^{-4}$, and a mini-batch size of 256. The learning rate is cosine-decayed to zero over the course of training. Both the VQ-VAE and DFoT use GELU activations and LayerNorm in their feed-forward stacks, and DFoT relies on the standard post-norm Transformer encoder architecture. For DFoT, we adopt a standard discrete diffusion formulation with an $x_0$-prediction parameterization. During training, we sample a diffusion timestep $t \in \{1, \dots, 1000\}$ independently for each token and train the model to predict the corresponding clean motion $x_0$. At inference time, we use DDIM sampling with 30 steps. Unless otherwise stated, we train each VQ-VAE model for 100,000 epochs and each DFoT model for 300,000 epochs, and select the checkpoint with the lowest validation loss. The network architecture, including the hidden dimensions, number of layers, and codebook size of the VQ-VAE, is summarized in Table~\ref{tab:net_details}.

All experiments are conducted on a single NVIDIA RTX A6000 GPU with 48\,GB of memory. A full DFoT model with the configuration in Table~\ref{tab:net_details} requires approximately 1 day of training time.

\begin{table*}[t]
  \centering
  \scriptsize
  \begin{tabular}{l l l l l}
    \toprule
    \textbf{Module} & \textbf{Component} & \textbf{Input shape} & \textbf{Operation} & \textbf{Hyperparameters} \\
    \midrule
    \multirow{3}{*}{\textbf{VQ-VAE}}
      & Encoder 
      & $(T, P, D_\text{in} + D_\text{cond})$ 
      & 1D Conv + ResNet stack 
      & 2 layers, hidden dim $d_\text{vq} = 512$ \\
      & Codebook 
      & $(T', P, d_\text{vq})$ 
      & Vector quantization 
      & $K = 1024$ codes, dim $d_\text{vq}$ \\
      & Decoder 
      & $(T', P, d_\text{vq} + D_\text{cond})$ 
      & Mirror of encoder 
      & 2 layers, hidden dim $d_\text{vq}$ \\
    \midrule
    \multirow{6}{*}{\textbf{DFoT}}
      & Noise emb.
      & $(T', P, d_\text{emb})$
      & Sinusoidal embedding
      & $d_\text{emb} = 256$ \\
      & Agent emb.
      & $(T', P, d_\text{emb})$
      & Learned embedding
      & $d_\text{emb} = 256$ \\
      & Input proj.
      & $(T', P, D_\text{m} + 2d_\text{emb})$
      & Linear + LayerNorm
      & 3 layers, $D_\text{m} + 2d_\text{emb} \rightarrow d = 512$ \\
      & Time emb.
      & $(T', P, d)$
      & RoPE
      & Applied to hidden dim $d$ \\
      & Core blocks 
      & $(T' \times P, d)$ 
      & Transformer blocks 
      & 6 layers, 8 heads, MLP dim $4d$ \\
      & Output head 
      & $(T', P, d)$ 
      & Linear projection 
      & $d \rightarrow D_\text{m}$ \\
    \bottomrule
  \end{tabular}
  \caption{\textbf{Network configuration.} $T$ is the input sequence length where $T' = T/\omega$ is the number of DFoT tokens ($\omega=4$) after temporal compression, $P$ the number of agents, $D_\text{in}$ the per-agent pose dimension, $D_\text{cond}$ the conditioning dimension, and $D_\text{m} = d_\text{vq} + 4 \times 9 \times P$ the full motion token dimension including pairwise transforms.}
  \label{tab:net_details}
\end{table*}

\section{Implementation Detail}
\label{sec:impl_supp}

\subsection{DFoT}
To train DFoT effectively for multi-agent motion generation, the raw motion data underwent a series of standardization and augmentation steps. First, all motion sequences were temporally harmonized by uniformly downsampling to 30 fps and spatially projected onto the xz-plane to ensure consistent ground-level representation. To enhance generalization, we applied mirror augmentation and randomly shuffled person identities during training, preventing the model from overfitting to specific individuals or movement directions. Finally, all processed features were standardized using z-score normalization before being fed into the DFoT model. We plan to opensource our code upon publication.

\subsection{Inference Speed}
Table~\ref{tab:fps_generator} demonstrates that \ourmethod{} runs at up to 56 FPS, this means \ourmethod{} can generate one future motion frame in under 18 ms for both partner and dyadic future prediction tasks, which significantly faster than competing methods. This inference speed experiment was conducted on a single NVIDIA RTX A6000 gpu with no smoothing guidance.  In the Partner Inpainting task, we achieve 54 FPS (vs. 49 FPS for Duolando and just 1 FPS for ReMoS). For Partner Prediction, our method provides a $\sim3.5 \times$ speedup over Ready-to-React (56 vs. 16 FPS), and for Dyadic future prediction we are $\sim 6.8 \times$  faster (54 vs. 8 FPS). These results highlight \ourmethod{}’s ability to deliver real-time performance while maintaining SoTA level motion quality and robust motion generation capabilities.

\begin{table}[h!]
\footnotesize
    \centering
    \begin{tabular}{l l c}
    \toprule
        Task & Method & FPS \\
    \midrule
        Partner Inpainting& Duolando & 49 fps\\
        & ReMoS & 1 fps\\
        & Ours& 54 fps\\
    \midrule
        Partner Prediction& Ready-to-React& 16 fps \\
         & Ours & 56 fps\\
    \midrule

 Dyadic Future Prediction& Ready-to-React& 8 fps\\
 & Ours& 54 fps\\

     \bottomrule

    \end{tabular}
    \caption{\textbf{Inference Speed Comparison}. Frames per second (FPS) comparison of \ourmethod{} against state-of-the-art baselines, demonstrating superior real-time performance.}
    \label{tab:fps_generator}
\end{table}

\begin{figure*}[htpb]
    \centering
    \includegraphics[width=0.8\linewidth]{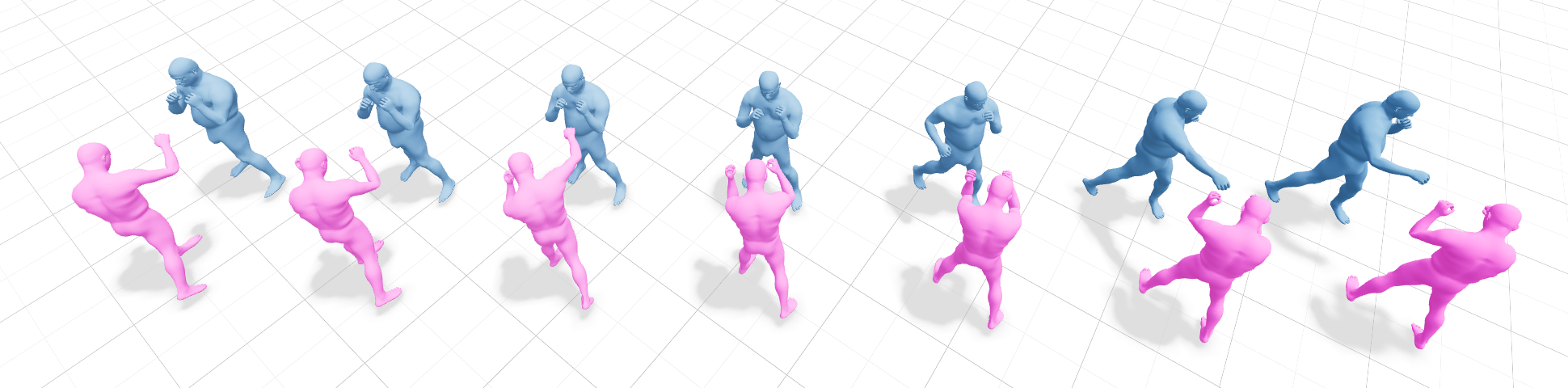}
    \caption{\textbf{Motion Control}. Agent A (pink) serves as the motion controller of Agent B (blue), with Agent B's next action predicted from Agent A's current and historical motion. This interaction shows adaptive responsive and coordination of Agent B as Agent A initiates an attack, prompting Agent B to block and counter.}
    \label{fig:motion_control}
\end{figure*}

\section{Joint Generation of Social Interaction}
We evaluate \ourmethod{} on the task of joint future prediction across two social interaction datasets: InterX~\cite{xu2024inter}, a dyadic interaction dataset, and Embody3D~\cite{mclean2025embody}, which contains scenes with 2--4 interacting people. For Embody3D, we train a single unified model on the full multi-person training set and evaluate separately on the 2-, 3-, and 4-person subsets, demonstrating our framework's ability to handle a variable number of agents without requiring separate models. Results are reported in \cref{tab:social_interaction}. On InterX, \ourmethod{} achieves strong generation quality with an FID of 0.210 and diverse outputs (DIV of 2.911), while maintaining low reconstruction errors (MPJPE of 0.475, MPJVE of 0.013). On Embody3D, the model generalizes well across varying group sizes. Notably, the 3-person subset yields the lowest FID (0.293) and MPJPE (0.526), suggesting that the model captures group dynamics most effectively at this scale, possibly due to a favorable balance of interaction complexity and data availability. The 2- and 4-person subsets exhibit higher FID scores (0.298 and 0.293, respectively), which may reflect greater distributional variability in dyadic scenes and increased combinatorial complexity in 4-person interactions. Across all subsets, the model maintains consistently low MPJVE values ($\leq$0.009), indicating temporally smooth and physically plausible generated motions. These results highlight the flexibility of our unified multi-agent formulation, which accommodates varying numbers of interacting agents within a single model while producing high-quality, diverse, and physically coherent social interactions.

\begin{table}[t]
\centering
\scriptsize
\setlength{\tabcolsep}{3pt}
\begin{tabular}{cl ccccccc}
\toprule
\multirow{2}{*}{\textbf{Dataset}}
& & \multicolumn{7}{c}{\textbf{Joint Future Prediction}} \\
\cmidrule(lr){3-9}
& & \textbf{FID} & \textbf{DIV} & \textbf{MI} & \textbf{FS} & \textbf{IP} & \textbf{MPJPE} & \textbf{MPJVE} \\
\midrule
InterX     &2 people & 0.210 & 2.911 & 0.130 & 0.074 & 0.093 & 0.475 & 0.013 \\
\midrule
\multirow{3}{*}{Embody3D}
& 2 people  & 0.298& 6.219& 0.093& 0.207& 0.023& 0.561& 0.011 \\
& 3 people  & 0.293& 6.313& 0.004& 0.257& 0.043& 0.526& 0.010 \\
& 4 people  & 0.590& 3.018& 0.058& 0.237& 0.050& 0.546& 0.009\\
\bottomrule
\end{tabular}
\caption{\textbf{Evaluation on Social Interaction Dataset.} Our model is evaluated on joint future prediction on InterX and Embody3D. For Embody3D, we train a single model on the full multi-person set (2--4 people) and report metrics separately on the 2-, 3-, and 4-person subsets.}
\vspace{-0.5em}
\label{tab:social_interaction}
\end{table}

\section{Additional Ablations} 

\subsection{Design Choice Ablations} 
Table~\ref{tab:method_comparison} ablates our positional encoding design. Agent PE informs the model \emph{which} agent a token belongs to, while Time PE encodes \emph{when} it occurs in the sequence. Removing Agent PE degrades FD by $11\times$ (0.560 vs.\ 0.052) and foot skating by $2.5\times$, indicating the model can no longer distinguish between agents and produces spatially incoherent motion.  Removing Time PE has an even larger impact on motion quality  (FD: 2.032), as the model loses temporal ordering  and generates implausible transitions. Removing both yields the worst velocity estimation (MPJVE: 0.094)  despite artificially high diversity, suggesting the model collapses to producing varied but unstructured motion. Our full model achieves the best scores across all metrics except diversity, which we attribute to the stronger coordination constraints learned when both embeddings are present.

\begin{table}[ht]
\centering
\vspace{-5pt}
\scriptsize
\setlength{\tabcolsep}{3pt}
\resizebox{\linewidth}{!}{%
\begin{tabular}{l ccccccc}
\toprule
\textbf{Variant} & FD$\downarrow$ & DIV$\uparrow$ & MI$\downarrow$ & FS$\downarrow$ & IP$\downarrow$ & MPJPE$\downarrow$ & MPJVE$\downarrow$ \\
\midrule
w/o Agent PE         & 0.560 & 14.029 & 0.148 & 1.078 & 0.382 & 0.810 & 0.075 \\
w/o Time PE          & 2.032 & 15.062 & 0.365 & 0.522 & 0.363 & 0.828 & 0.061 \\
w/o Agent \& Time PE & 0.920 & \textbf{15.081} & \textbf{0.032} & 0.974 & 0.370 & 0.835 & 0.094 \\
\midrule
\textbf{Full model (Ours)} & \textbf{0.052} & 9.572 & 0.124 & \textbf{0.423} & \textbf{0.116} & \textbf{0.641} & \textbf{0.034} \\
\bottomrule
\end{tabular}%
}
\caption{\textbf{Ablation of positional encodings.} Removing Agent or Time PE degrades motion quality (FD) by $11\times$ and $39\times$ respectively. Our full model achieves the best quality and coordination across all metrics. The higher DIV in ablated variants reflects unstructured motion rather than meaningful diversity.}
\label{tab:method_comparison}
\end{table}

\subsection{Noise Sampling Strategy Ablations}
Table~\ref{tab:noise_sampling} compares two noise sampling strategies for the joint future prediction task on DuoBox. In \emph{Joint Sampling}, both agents at each timestep share the same noise level, preserving spatiotemporal correlations during denoising. In \emph{Independent Sampling}, each agent's token receives an independently sampled noise level. Joint Sampling substantially outperforms Independent Sampling on FD ($0.052$ vs.\ $0.716$), diversity ($9.572$ vs.\ $1.804$), and interaction quality (IP: $0.116$ vs.\ $0.231$). Independent Sampling achieves lower foot skating and MPJPE, suggesting it produces kinematically conservative motion---but at the cost of diversity and expressivity. We attribute this to the decorrelated noise breaking the joint structure between agents: each agent denoises plausible but overly safe motion independently, resulting in muted interactions. We use Joint Sampling for all joint prediction experiments.

\begin{table}[ht]
\centering
\vspace{-5pt}
\footnotesize
\setlength{\tabcolsep}{3pt}
\resizebox{\columnwidth}{!}{
\begin{tabular}{l ccccccc}
\toprule
\textbf{Noise Sampling} & FD$\downarrow$ & DIV$\uparrow$ & MI$\downarrow$ & FS$\downarrow$ & IP$\downarrow$ & MPJPE$\downarrow$ & MPJVE$\downarrow$ \\
\midrule
Independent (per-token) & 0.716 & 1.804 & 0.155 & \textbf{0.223} & 0.231 & \textbf{0.608} & 0.021 \\
\textbf{Joint (Ours)} & \textbf{0.052} & \textbf{9.572} & \textbf{0.124} & 0.423 & \textbf{0.116} & 0.641 & \textbf{0.006} \\
\bottomrule
\end{tabular}
}
\caption{\textbf{Joint vs.\ independent noise sampling on DuoBox.} Joint sampling preserves inter-agent correlations during denoising, yielding substantially better distribution quality (FD) and diversity while maintaining strong coordination. Independent sampling produces kinematically conservative but less expressive motion.}
\label{tab:noise_sampling}
\end{table}

\section{Motion History Guidance}
Drawing inspiration from History Guidance, we present Motion History Guidance for controllable multi-agent motion generation. Our method decomposes the guidance signal into specific historical dependencies. Specifically, we compute $\mathbf{M}_\text{cond}$ using the full history, $\mathbf{M}_\text{uncond}$ without any history, and two specialized terms: $\mathbf{M}_\text{self}$, which captures individual motion continuity, and $\mathbf{M}_\text{partner}$, which focuses on social interactions by conditioning only on other agents' histories. The terms are defined as: 

\begin{align*}
\mathbf{M}_\text{cond} =& f_{\phi}(\mathbf{M}^\text{agent1}_{t:L}, \mathbf{M}^\text{agent2}_{t:L}, ... \,|\,\mathbf{M}^\text{agent1}_{0:t-
1}, \mathbf{M}^\text{agent2}_{0:t-1}, ...) \\
\mathbf{M}_\text{uncond} =& f_{\phi}(\mathbf{M}^\text{agent1}_{t:L}, \mathbf{M}^\text{agent2}_{t:L}, ... \,|\,\text{unconditional}) \\
\mathbf{M}_\text{self} =& \frac{1}{N}\sum_{n}^{N} f_{\phi}(\mathbf{M}^\text{agent1}_{t:L}, \mathbf{M}^\text{agent2}_{t:L}, ... \,|\,\mathbf{M}^\text{agent\,n}_{0:t-1}) \\
\mathbf{M}_\text{partner} =&\ \frac{1}{N}\sum_{n=1}^{N} f_{\phi}(\mathbf{M}^\text{agent1}_{t:L}, \mathbf{M}^\text{agent2}_{t:L}, \ldots \,|\, \{\mathbf{M}^\text{agent\,k}_{0:t-1}\}_{k\neq n})
\end{align*}

Using these components, we formulate two history guidance (HG) variants, self history guidance (SHG) and partner history guidance (PHG), to steer the generation process:

\begin{align*}
\textbf{HG:} \quad 
& \mathbf{M} = (1+w)\mathbf{M}_\text{cond} - w \mathbf{M}_\text{uncond} \\[6pt]
\textbf{SHG:} \quad 
& \mathbf{M} =  \mathbf{M}_\text{cond} + w \mathbf{M}_\text{self} - w \mathbf{M}_\text{uncond} \\[6pt]
\textbf{PHG:} \quad 
& \mathbf{M} = \mathbf{M}_\text{cond} + w \mathbf{M}_\text{partner} - w \mathbf{M}_\text{uncond} \\[6pt]
\end{align*}

Here, $w$ is the guidance weight, controlling the strength of the target historical influence relative to the unconditional prediction; in all experiments, we set $w=1$.

\section{Smoothing Guidance}

During inference the denoiser $f_{\boldsymbol{\phi}}$ iteratively refines the noisy token sequence $\mathbf{M}^{(k)}$, producing at each step a clean-motion estimate $\hat{\mathbf{M}}_{\mathbf{0}} = f_{\boldsymbol{\phi}}(\mathbf{M}^{(k)}, \boldsymbol{k}_{\text{seq}})$. Each denoising step sees only the current noisy state and predicts the clean tokens independently---nothing forces adjacent token timesteps to agree on velocity or acceleration. The result can be a motion that is plausible per-frame but jittery when played back.

\paragraph{Smoothness objective.}
We inject a preference for temporal coherence directly into the sampling process. At each denoising step we inspect the predicted clean sequence $\hat{\mathbf{M}}_{\mathbf{0}}$ for roughness via the discrete second-order finite difference (acceleration) along the token-time axis. For the $p$-th agent at token timestep~$i$:
\begin{multline}
    \Delta^{2}\hat{\mathbf{m}}^{p}_{0}[i] \;=\; \hat{\mathbf{m}}^{p}_{0}[i-1] \;-\; 2\,\hat{\mathbf{m}}^{p}_{0}[i] \;+\; \hat{\mathbf{m}}^{p}_{0}[i+1], \\
    i \in \{2,\dots,T'-1\},
    \label{eq:laplacian}
\end{multline}
where $T' = T/\omega$ is the number of token timesteps after VQ-VAE temporal compression. This quantity is zero for constant-velocity motion and large wherever the trajectory has kinks. We aggregate it into a scalar roughness objective over all $P$ agents:
\begin{equation}
    \mathcal{L}_{\mathrm{smooth}} \;=\; \sum_{p=1}^{P}\sum_{i=2}^{T'-1} \left\|\, \Delta^{2}\hat{\mathbf{m}}^{p}_{0}[i] \,\right\|^{2}.
    \label{eq:smooth_loss}
\end{equation}

\paragraph{Guided sampling step.}
The gradient $\nabla_{\mathbf{M}^{(k)}}\mathcal{L}_{\mathrm{smooth}}$ indicates the direction in noise-space that would most reduce the predicted roughness. We obtain it by backpropagating Eq.~\eqref{eq:smooth_loss} through the denoiser $f_{\boldsymbol{\phi}}$ and modify the standard denoising update:
\begin{equation}
    \mathbf{M}^{(k)} \;\leftarrow\; \mathbf{M}^{(k)} \;-\; \lambda\,\nabla_{\mathbf{M}^{(k)}}\mathcal{L}_{\mathrm{smooth}},
    \label{eq:guided_update}
\end{equation}
where $\lambda$ controls the guidance strength. This steers the sampling trajectory toward solutions that are not only plausible under the learned distribution but also temporally coherent, without any retraining.

\paragraph{Relation to classifier guidance.}
This follows the same principle as classifier guidance in image diffusion~\cite{dhariwal2021diffusion}: an external signal---there a classifier, here a hand-crafted smoothness objective---steers generation at inference time via $\nabla_{\mathbf{M}^{(k)}}$ without modifying the pretrained denoiser $f_{\boldsymbol{\phi}}$. A single model produces either raw samples or temporally smooth ones, controlled entirely by~$\lambda$.

\begin{figure*}[t]
    \centering
    \includegraphics[width=1\linewidth]{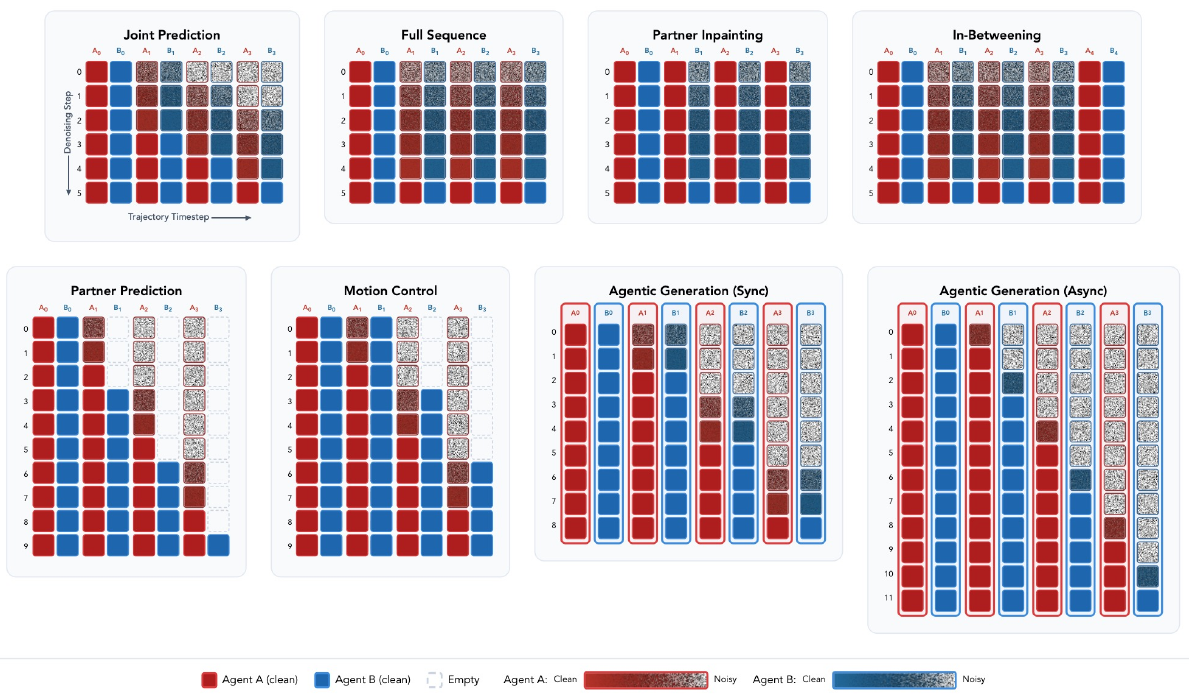}
    \caption{Sampling strategies supported by a single \ourmethod{} model at inference time. The top row shows joint prediction, full sequence denoising, partner inpainting, and in-betweening. The bottom row shows partner prediction, motion control, agentic generation (synchronous), and agentic generation (asynchronous). Red and blue represent Agent A and B, with saturation indicating noise level across denoising steps. Dashed outlines mark empty (unavailable) token positions.}
    \label{fig:sampling_strategies}
\end{figure*}

\section{Other Sampling Strategies}
Additional sampling strategies not discussed in the main paper include In-Betweening and Motion Control, which we describe below. Figure \ref{fig:sampling_strategies} illustrates all the sampling strategies we have explored.

\textbf{In-Betweening.}
Given an arbitrary set of predefined keyframes for both agents, the goal of the in-betweening task is to generate the continuous motion between the predefined keyframes. Let $\mathcal{T}$ denote the discrete set of the predefined keyframes, and let
\begin{equation}
\mathcal{G} = \{t \mid \, t \notin \mathcal{T}\}
\end{equation}
be the set of frames to be generated. The objective of the in-betweening task is thus defined as
\begin{equation}
P(A_{\mathcal{G}}, B_{\mathcal{G}} \mid A_{\mathcal{T}}, B_{\mathcal{T}})
\end{equation}

where the model generates motions only for the non-keyframe $\mathcal{G}$ while strictly adhering to the keyframes in $\mathcal{T}$. See video teaser for results.

This formulation ensures that all generated frames are distinct from the given keyframes and promotes smooth, temporally coherent, and coordinated transitions between them.

\textbf{Motion Control.} In the motion control task, Agent A’s next action ($A_t$) is predicted based on all of its past motion and the partner’s current motion ($B_t$), enabling direct and adaptive control of Agent A’s behavior in response to the partner’s movements:
\begin{equation}
P(A_{t} \mid A_{0:t-1}, B_{0:t})
\end{equation}
This formulation allows Agent B to serve as a motion controller of Agent A's motion, facilitating responsive and coordinated interaction (see Fig~\ref{fig:motion_control}).

\end{document}